%% file: main.tex
\documentclass[11pt]{article}

\usepackage[final]{acl}

\usepackage{times}
\usepackage{latexsym}

\usepackage[T1]{fontenc}
\usepackage[utf8]{inputenc}

\usepackage{microtype}

\usepackage{inconsolata}
\usepackage{microtype}
\usepackage{graphicx}
\usepackage{subfigure}
\usepackage{booktabs} % for professional tables

\usepackage{float}
\usepackage{makecell,multirow,diagbox}  
\usepackage{float}
\usepackage{adjustbox}
\usepackage{lscape}
\usepackage{amsfonts}
\usepackage{amsthm}
\usepackage{amsmath}

\usepackage{amssymb}
\usepackage{color}
\usepackage{xcolor}
\usepackage{algorithm}
\usepackage[noend]{algorithmic}
\usepackage{enumitem}
\usepackage{color, colortbl}\usepackage{url}
\definecolor{gray}{HTML}{545454}%black!15,545454,2F4F4F
\definecolor{bluee}{HTML}{77CDFF}
\usepackage{booktabs}
\usepackage{multirow}
\usepackage[capitalize,noabbrev]{cleveref}
\usepackage[textsize=tiny]{todonotes}
\usepackage{amsmath}
\usepackage{amsfonts}
\usepackage{utfsym}
\usepackage{tcolorbox}
\usepackage{pifont}
\usepackage{flushend}

\input{preamble_math}
\title{Enhancing Tool Learning in Large Language Models with\\ Hierarchical Error Checklists}

\author{Yue Cui$^{1,2}$, Liuyi Yao$^{1}$, Shuchang Tao$^{1}$, Weijie Shi$^{2}$, Yaliang Li$^{1}$ \\
\textbf{Bolin Ding$^{1}$, and Xiaofang Zhou$^{2}$} \\
$^1$Alibaba Group\\
$^2$The Hong Kong University of Science and Technology \\ 
\texttt{\{ycuias,ashiah,zxf\}@cse.ust.hk,} \\ 
\texttt{\{yly287738,taoshuchang.tsc,yaliang.li,bolin.ding\}@alibaba-inc.com} \\}

\begin{document}
\maketitle
\begin{abstract}
Large language models (LLMs) have significantly advanced natural language processing, particularly through the integration of external tools and APIs. However, their effectiveness is frequently hampered by parameter mis-filling during tool (function) calling. In this paper, we propose the Hierarchical Tool Error Checklist (HiTEC) framework to systematically diagnose and mitigate tool-calling errors without relying on extensive real-world interactions. HiTEC introduces a two-tiered approach: a global error checklist that identifies common, cross-tool issues, and a local error checklist that targets tool-specific and contextual failures. Building on this structure, we propose two deployments: HiTEC-In Context Learning (HiTEC-ICL) and HiTEC-Kahneman-Tversky Optimization (HiTEC-KTO). HiTEC-ICL embeds the global checklist in the initial prompts and leverages a two-round conversational interaction to dynamically refine parameter handling, while HiTEC-KTO generates high-quality negative examples to drive fine-tuning via preference-based optimization. Extensive experiments conducted on five public datasets show that our framework improves parameter-filling accuracy by up to 42\% compared to baseline methods. 
%Code is available at \url{https://anonymous.4open.science/status/hitec-E573}.
\end{abstract}

\input{subfiles/1_Intro}

%\input{subfiles/2_Pre}
\input{subfiles/5_Related}

\input{subfiles/3_Method}
\input{subfiles/4_Experiment}
\input{subfiles/6_Conclusion}

\section*{Acknowledgments} 
The research work described in this paper was supported by Hong Kong Research Grants Council (grant\# 16202722, T22-607/24-N, T43-513/23N-1). It was partially conducted in JC STEM Lab of Data Science Foundations funded by The Hong Kong Jockey Club Charities Trust.

\clearpage
\section*{Limitation}
While our approach shows promising improvements, it has some limitations. The effectiveness of HiTEC partially attributes to the comprehensiveness of the manually crafted error types and templates, which may not capture all novel or unforeseen errors in dynamic tool environments. Additionally, the reliance on simulated error feedback may not fully reflect real-world scenarios, potentially limiting the framework's generalizability and scalability in diverse applications. 

To address these limitations, potential improvements could focus on dynamically updating the error checklists based on real-world feedback. By integrating mechanisms for continuous monitoring and analysis of actual tool interactions, the system could iteratively refine the types of errors included in the checklists. This iterative improvement would allow the framework to adapt to evolving tool behaviors and emerging error patterns, ultimately enhancing its robustness and precision in tool calling. Results in Section \ref{sec:error_type} can also verify effectiveness of such an extension. Additionally, incorporating automated error detection and leveraging reinforcement or continual learning strategies may further optimize the system's performance in diverse and dynamic environments.

%\balance
\bibliography{refs}

\input{subfiles/7_Appendix}
\end{document}

%% file: preamble_math.tex
\newtheorem*{theorem*}{Theorem}
\newtheorem*{definition*}{Definition}
\newtheorem*{assumption*}{Assumption}
\newtheorem*{conjecture*}{Conjecture}
\newtheorem*{claim*}{Claim}
\newtheorem*{lemma*}{Lemma}
\newtheorem*{proposition*}{Proposition}
\newtheorem*{property*}{Property}
\newtheorem*{fact*}{Fact}
\newtheorem*{corollary*}{Corollary}
\newtheorem*{example*}{Example}
\newtheorem*{remark*}{Remark}
\newtheorem*{exercise*}{Exercise}

%% file: subfiles/1_Intro.tex
% !TEX root = ../main.tex

\section{Introduction}

Large language models (LLMs) have revolutionized natural language processing by enabling advanced comprehension, reasoning, and task execution capabilities. Among these, the integration of external tools and application programming interfaces (APIs) represents a critical milestone, allowing LLMs to expand their utility beyond textual analysis into interactive, task-oriented domains \cite{qu2025tool,li2023api,qintoolllm,wu2024seal,tang2023toolalpaca,song2023restgpt,gou2023critic}. To achieve this, LLMs must navigate a complex process of function calling, which involves selecting the appropriate tools, formulating precise input arguments, and parsing results to satisfy tool input. Despite these advancements, tool-calling often suffers from incorrect parameter filling \cite{lin2024hammer,liu2024tool}, a recurring issue that undermines the accuracy and reliability of LLM-driven interactions. 
\begin{figure*}
    \vspace{-0.45cm}
    \centering
    \includegraphics[width=\linewidth]{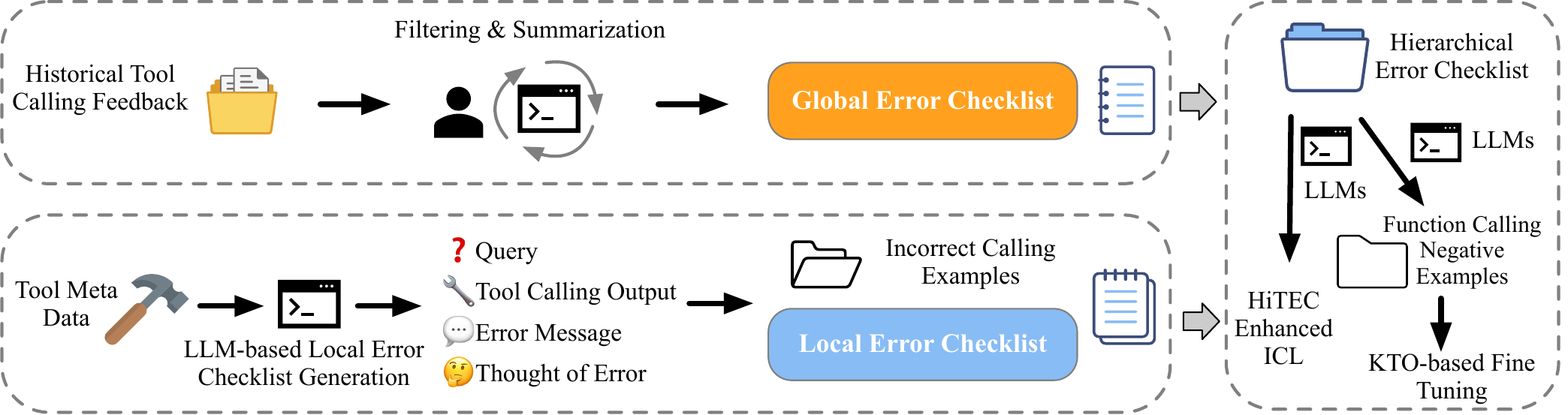}
    \vspace{-0.3cm}
    \caption{Pipeline of the Proposed Hierarchical Tool Error Checklist (HiTEC), which includes Global Error Checklist and Local Error Checklist}
    \vspace{-0.3cm}
    \label{fig:enter-label}
    %\vspace{-0.3cm}
\end{figure*}

Most previous tool learning methods require LLM-tool interactions to improve the calling accuracy \cite{wang2024llms,zhang2023reverse,shi2024learning,yao2022react,qintoolllm,chen2024advancing,yang2024gpt4tools}. For example, STE \citep{wang2024llms} simulate plausible scenarios and incorporates execution feedback to enhance the correct use of tools. It involves first simulating queries, executing real tool calls via tool-LLM interactions, and learning from function calling outputs when errors occur. While real-world interactions with tools can yield valuable insights, they cause intensive resources (For example, 10-25\$/1,000 transactions for Bing Search API \footnote{\url{https://www.microsoft.com/en-us/bing/apis/pricing}}) and instability issues \citep{guo2024stabletoolbench}. Furthermore, the errors encountered by most tools called by LLMs are predominantly common types that can be known in advance before real tool calling. To address these challenges, we argue that (1) tool-LLM interactions that involve real-world tool callings are not always necessary, and (2) structured error checklists can systematically identify potential performance deficiencies and guide targeted improvements.

Based on this motivation, this paper proposes a Hierarchical Tool Error Checklist (HiTEC) framework to enhance tool learning with LLMs by identifying and addressing tool-calling errors. As shown in Figure \ref{fig:enter-label}, The HiTEC framework is composed of the constructions of two levels of error checklists: a global error checklist, which captures general errors that frequently occur across different tools, and a local error checklist, which focuses on tool-specific errors and contextual failures. These checklists provide a structured and comprehensive way to diagnose and rectify tool-calling errors without requiring extensive real-world execution. 

We then deploy HiTEC in tuning-free and tuning-based ways and propose: HiTEC-In Context Learning (HiTEC-ICL) and HiTEC-Kahneman-Tversky Optimization (HiTEC-KTO). By embedding the global error checklist in the initial query and integrating the local error checklist through a two-round conversational interaction, HiTEC-ICL guides LLMs to preemptively avoid common mistakes and refine parameter handling flexibly. HiTEC-KTO leverages the error checklists to generate high-quality negative examples, which are then used to empower open-source LLMs through KTO-based fine-tuning. This strategy equips models with enhanced function calling accuracy by training them to recognize and correct errors. Our main contributions are as follows:

% This paper makes the following key contributions:

\scalebox{0.6}{\ding{108}} \emph{Framework.} We propose a novel hierarchical error checklist framework that integrates global and local error messages to enhance tool learning of LLMs. The framework distinguishes between offline self-generated errors and online tool-LLM interaction based accumulated errors to support adaptive and scalable tool learning.

\scalebox{0.6}{\ding{108}} \emph{Method.} We propose two novel methods: (1) HiTEC-ICL, an efficient mechanism to integrate error messages into LLM prompts for dynamic error reflection and correction; and (2) HiTEC-KTO, a negative sample generation method that overcomes the failure modes of preference-based optimization through fine-tuning with negative examples. 

\scalebox{0.6}{\ding{108}} \emph{Analysis.} We theoretically and empirically demonstrate the effectiveness of KTO-based tuning in overcoming the failure modes of preference-based optimization in tool learning, thereby enabling LLMs to enhance their tool-calling accuracy through fine-tuning with negative examples.

\scalebox{0.6}{\ding{108}} \emph{Performance.} We conduct extensive experiments across five public datasets. Our results demonstrate improvements in parameter-filling accuracy, tool-calling success rates compared to baseline methods.

%% file: subfiles/5_Related.tex
% !TEX root = ../main.tex

\section{Related Work}

LLM-based tool learning methods can be generally categorized into tuning-free and tuning-based approaches \cite{qu2025tool}. Tuning-free methods leverage the inherent capabilities of large language models by prompting them to interact directly with external tools. This category includes techniques such as few-shot demonstrations \cite{wang2024llms}, rule-based strategies \cite{zhang2023reverse,shi2024learning,yao2022react}, and the use of optimized tool descriptions \cite{yuan2024easytool,chen2024re} to facilitate efficient parameter extraction and tool usage. Owing to their simplicity and scalability, tuning-free methods have gained popularity; however, their performance often lags behind that of tuning-based approaches and a finally correct tool calling may require multiple times of tool-LLM interactions, which can be costly.

\begin{figure*}[h!]
    \vspace{-0.45cm}
    \centering
    \includegraphics[width=0.95\linewidth]{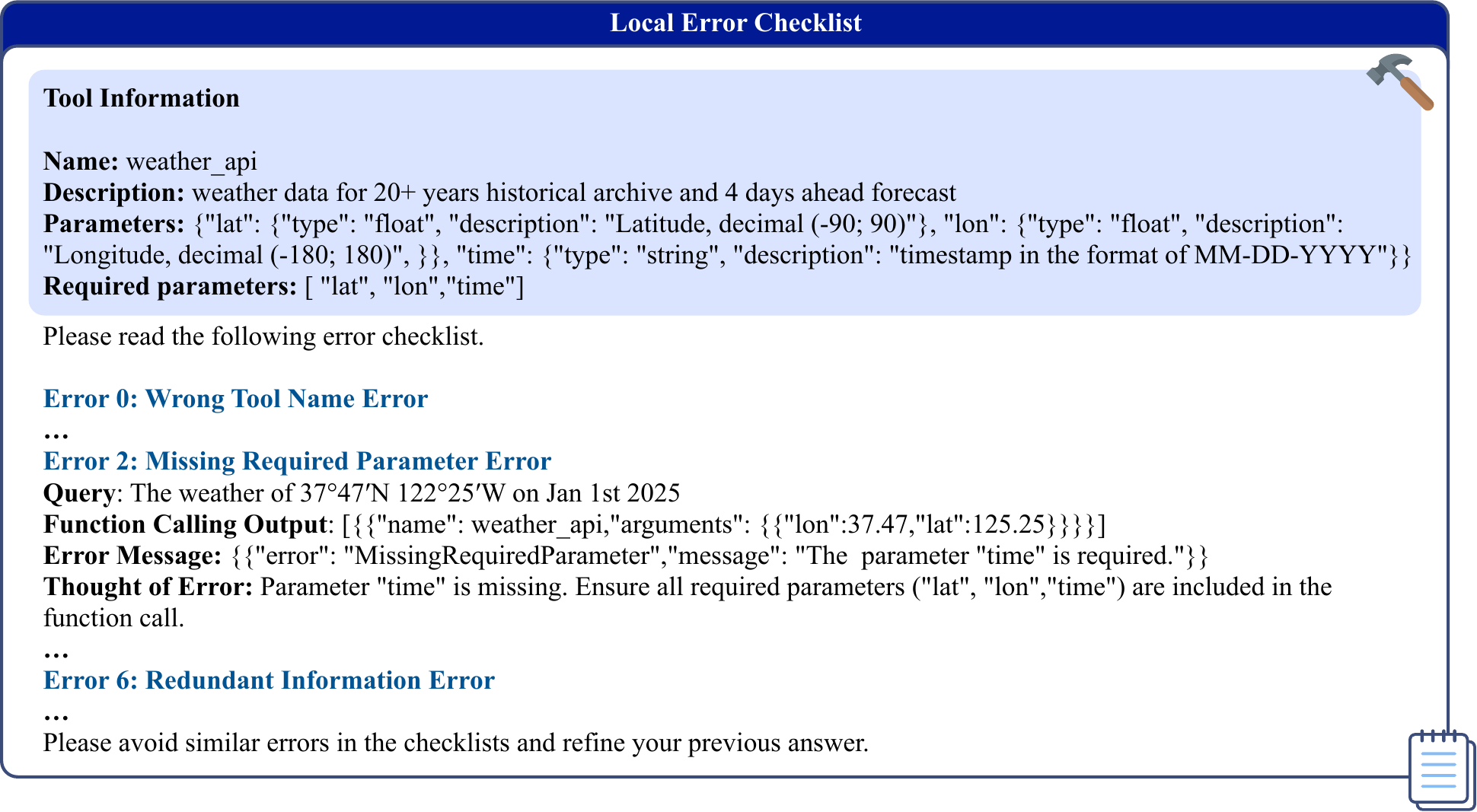}
    \vspace{-0.3cm}
    \caption{The Local Error Checklist: a list of tool-specific issues that may arise during tool calling}
    \label{fig:local_ec}
    % \vspace{-0.45cm}
    \vspace{-0.2cm}
\end{figure*}

\begin{figure}[t]
    \vspace{-0.3cm}
    \centering
    \includegraphics[width=0.95\linewidth]{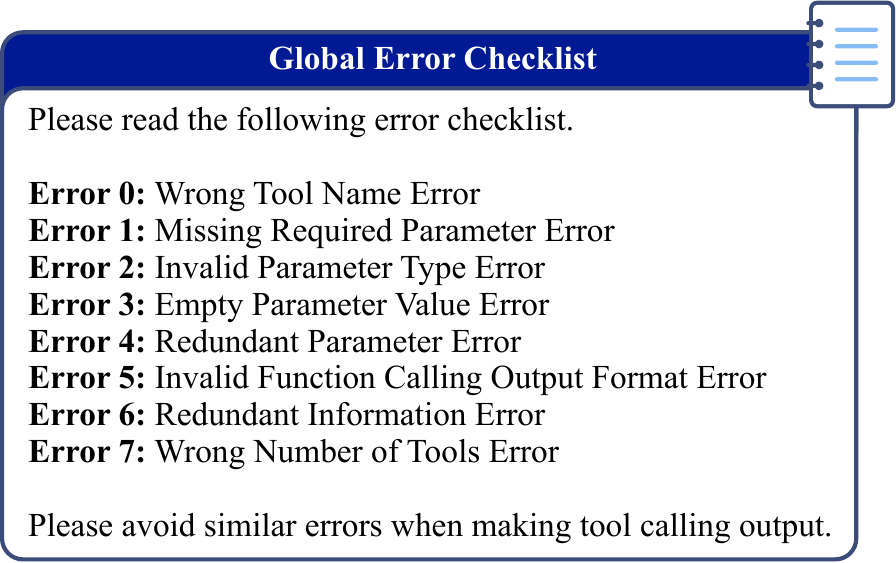}
    \vspace{-0.1cm}
    \caption{The Global Error Checklist: a list of common issues that may arise during tool calling}
    \label{fig:global_ec}
    \vspace{-0.2cm}
\end{figure}

In contrast, tuning-based methods enhance tool calling performance through fine-tuning strategies. GPT4Tools \cite{yang2024gpt4tools} uses LoRA-based supervised fine-tuning, while the introduction of tool-specific tokens is demonstrated in Toolkengpt \cite{hao2023toolkengpt}. Other approaches leverage augmented datasets \cite{lin2024hammer}, and utilize interactive path-based reasoning \cite{qintoolllm,chen2024advancing} to improve tool interaction precision. Relign \cite{xu2024reducing} tackles tool hallucinations (also incorrect tool selection/usage) by expanding the action space with "indecisive actions" (e.g., deferring tool use) and aligning reliability via preference optimization. Although tuning-based methods consistently deliver superior performance by tailoring LLMs specifically for tool calling tasks, they require high-quality training data or extensive tool interaction logs, which are still costly to obtain.

%% file: subfiles/3_Method.tex
% !TEX root = ../main.tex

\section{Method}
In this section, we introduce the novel Hierarchical Tool Error Checklist (HiTEC) framework, comprising global and local error checklists to address parameter mis-filling in tool learning. We propose both tuning-free (HiTEC-ICL) and tuning-based (HiTEC-KTO) methods to integrate HiTEC into LLMs, enabling dynamic error reflection and correction based on pre-identified patterns and tool-specific information. 
\subsection{Tool Error Checklist}
\subsubsection{Global Error Checklist}
The global error checklist is a list of common issues that arise during tool calling, and they are not associated with a specific query or tool. It is designed to address the most prevalent and impactful issues encountered during tool calling. Drawing from prior experiences with model-tool interactions, eight errors are selected to represent typical failure modes occurring at various stages of tool interaction, encompassing both tool-level and parameter-level mistakes. 

At the tool level, we check errors \textit{Wrong Tool Name Error} (Error 0) and \textit{Wrong Number of Tools Error} (Error 7) highlight issues related to improper tool selection or usage. The parameter level encompasses a range of potential inaccuracies, including \textit{Missing Required Parameter Error} (Error 1), \textit{Invalid Parameter Type Error} (Error 2), and \textit{Empty Parameter Value Error} (Error 3), and \textit{Redundant Parameter Error} (Error 4). In addition, the global error checklist also incorporates errors related to redundancy. We check\textit{Redundant Information Error} (Error 6), which ensures that the output remains concise and relevant, and \textit{Invalid Function Calling Output Format Error} (Error 5) guards against syntactical inconsistencies that could disrupt downstream processes. This carefully constructed checklist of errors provides a robust foundation for systematic error identification and resolution in diverse tool-calling scenarios, significantly enhancing the reliability and efficiency of model-tool interactions. The global error checklist is presented in Figure \ref{fig:global_ec}.

\subsubsection{Local Error Checklist}

The local error checklist identifies errors related to the specific features of each tool. This checklist is crucial for addressing tool-specific issues, as it offers detailed information that goes beyond the general overview provided by the global error checklist. Unlike the global error checklist, which only lists an overview of common errors, the local checklist focuses on the unique functionalities, parameters, and requirements of each tool. The local error checklist is essential for addressing tool-specific issues that may not be captured by the global checklist. A tailored local checklist ensures these specificities are considered. 

The local error checklist of a tool contains the following components: tool information, simulated queries for each error type, the function calling output that can invoke the corresponding error, an error message describing the error, and a Thought of Error reflection that specifies how such an error can be corrected. An example of a local error checklist is illustrated in Figure \ref{fig:local_ec}.

To form such a local error checklist, it is required that tool-related information be known in advance. This includes details such as those outlined in the "Tool Information" section of the Error Checklist. This information should be appropriately placed within the <tool\_info> field of the generation prompt as presented in Appendix \ref{app:prompt}.

%It is designed that the local error checklist serves as a tailored and practical guide to parameter filling of tools. Identifying and resolving tool-specific parameter issues, ensuring precise and efficient tool usage.

\subsection{HiTEC-ICL: Enhancing LLM Tool Calling in Tuning-free Way }
% The designed global and local error checklists are then integrated into the LLM-based tool-calling conversation to ensure precise and reliable tool utilization. 
We integrate the designed global and local error checklists into the LLM-based tool-calling conversation to ensure precise and reliable tool utilization.

The global error checklist is embedded within the user's initial query at the outset of the inference process. This proactive integration helps preempt common issues, such as tool name misidentification or parameter omission. By implementing these error prevention mechanisms early in the process, the system significantly enhances the accuracy and reliability of the initial tool invocation.

\begin{figure}[t]
    \vspace{-0.3cm}
    \centering
    \includegraphics[width=\linewidth]{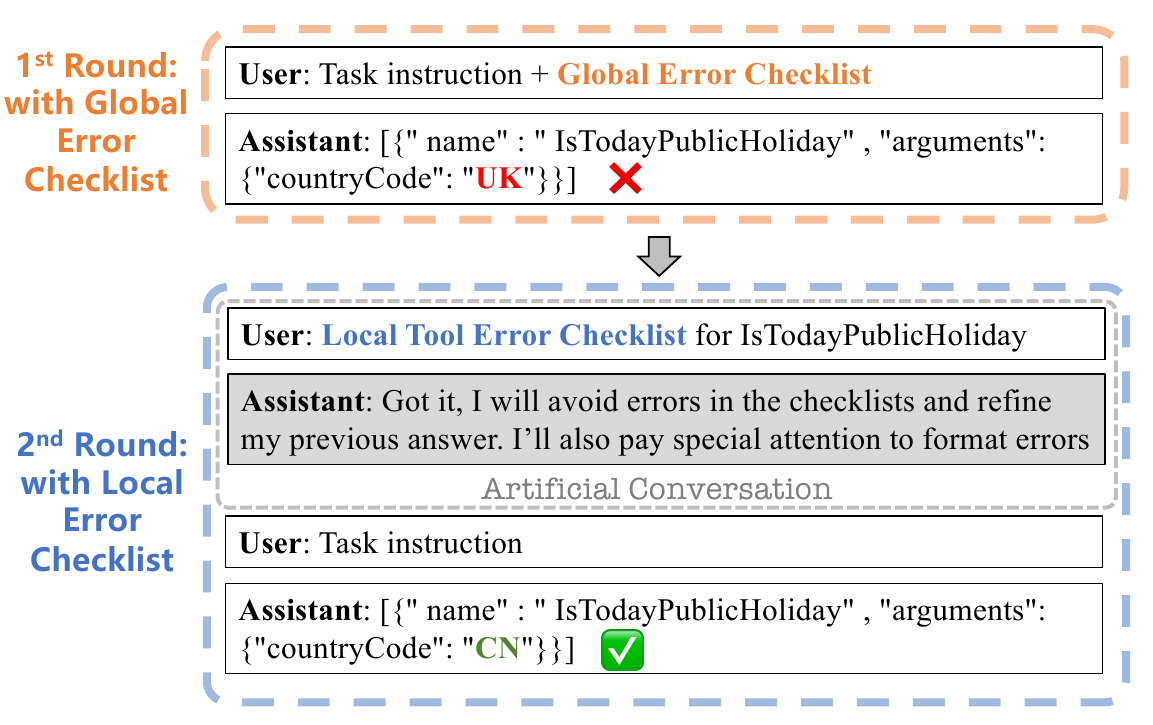}
    \vspace{-0.3cm}
    \caption{An Example Conversation with the Global-Local Checklist as ICL Prompting}
    \vspace{-0.3cm}
    \label{fig:glblc_conversation}
    %\vspace{-0.3cm}
\end{figure}

The local error checklist is primarily designed for parameter-specific error handling. Figure \ref{fig:glblc_conversation} illustrates this two-round interaction process. The first round captures the initial tool call, while the second round, guided by the local error checklist, ensures corrections are made to parameter filling and other tool-specific issues, leading to improved inference outcomes.

\subsection{HiTEC-KTO: Enhancing LLM Tool Calling in Tuning-based Way}

%liuyi Version 
To empower open-source language models with robust tool-calling capabilities, we propose a method that utilizes error checklists to generate high-quality negative examples. This approach aims to enhance the performance of smaller, efficient models, enabling them to rival larger, more resource-intensive counterparts. By curating negative examples based on predefined global and local error checklists, we fine-tune open-source models, thereby equipping them with improved function-calling accuracy and error-handling capabilities.

\subsubsection{Negative Example Generation} Negative examples are generated by combining correctly formatted tool-calling outputs with the corresponding local error checklist. By prompting the model to introduce specific errors described in the checklist, we create outputs that simulate real-world failures in tool-calling tasks. This curated dataset enables the fine-tuning of models to better recognize, avoid, and correct errors, ultimately improving their reliability and alignment with intended behaviors. We name the generated pairwise tool-calling dataset as PTC. The prompt for negative sample generation is presented in Appendix~\ref{app:prompt}.
\label{sec:negative_sample}

\subsubsection{Failure Mode of DPO on PTC Dataset} 
\label{sec:DPO-failure}
Given that the PTC dataset consists of pairwise data, an intuitive choice for tuning would be Direct Preference Optimization (DPO)~\cite{rafailov2024direct} as the tuning method. However, applying DPO to such a dataset leads to a failure mode, as identified in previous research~\cite{feng2024towards,pal2024smaug}. This issue arises because the positive and negative responses in the PTC dataset differ by only a few tokens (more details are in Appendix~\ref{appendix:dpo_kto}). Consequently, two problems occur: 1) the gradient of the DPO loss approaches zero, leading to a weak update signal, and 2) during optimization, the probability of the correct token tends to decrease when compared to the reference model (i.e., the initial model being tuned). 
We empirically validate the gradient vanish phenomenon by plotting the gradient norm during the training of DPO on the PTC dataset (for more details, refer to Figure~\ref{subfig:gradient_norm} in Appendix~\ref{appendix:dpo_kto}). Further, by plotting the log probabilities for both positive and negative samples (see Figure~\ref{subfig:logps_chosen} in Appendix~\ref{appendix:dpo_kto}), we observe that although the log probabilities of negative samples decrease, the log probabilities of positive samples also decrease, demonstrating the second challenge.

\subsubsection{HiTEC-KTO}
One representative DPO variant is Kahneman-Tversky Optimization (KTO)~\cite{ethayarajh2024kto}. We refer to the approach of fine-tuning large language models (LLMs) on a PTC-type dataset using KTO as HiTEC-KTO. In the following, we show the potential that KTO can address the above failure mode of DPO. 

% We begin by expressing the KTO loss in a paired format: 
We formulate the KTO loss in a paired format:
\begin{equation}
\begin{split}
    &\mathcal{L}_{KTO}(x, y_w, y_l; \theta) \\
    =& \lambda_w - \lambda_w\sigma\left(\beta(r_{\theta}(x, y_w) - z_0)\right)\\
    &+   \lambda_l - \lambda_l\sigma\left(\beta\left( z_0 - r_{\theta}(x, y_l)\right)\right),  
\end{split}
\label{eqn:kto-loss}
\end{equation}
where $(x, y_w, y_l)$ are paired data with $x$ as the prompt, $y_w$ as the positive answer and $y_l$ as the negative answer. 
$r_{\theta}(x, y)$ is the log-ratio of the likelihoods of answer $y$ between the training model $\pi_{\theta}(y|x)$ and the reference model $\pi_{\text{ref}}(y|x)$, where $\theta$ is the model parameter. $r_{\theta}(x, y)$ is denoted as $r_{\theta}(x, y) = \frac{\pi_{\theta}(y|x)}{\pi_{\text{ref}}(y|x)}$.
$z_0$ is the reference point, and $z_0 = KL(\pi_{\theta} || \pi_{\text{ref}})$. Since $z_0$ does not propagate gradients~\cite{ethayarajh2024kto}, it is treated as a constant during the differentiation of $\mathcal{L}_{KTO}$. 

Suppose $y_w$ and $y_l$ are length $K$ sequences, and they only differ at i-th token, i.e., $y_w= [t_1, \cdots, t_{i-1}, t_{i}^w, t_{i+1}, \cdots, t_K]$, and $y_l= [t_1, \cdots, t_{i-1}, t_{i}^l, t_{i+1}, \cdots, t_K]$.
Using the derivative perspective, similar to the approach in the prior section, we theoretically analyze KTO's capability to mitigate the failure mode of DPO. The derivative of the KTO loss with respect to $\theta$ is:
\begin{equation}
\small
\begin{split}
    &\nabla_{\theta}\mathcal{L}_{KTO}(x, y_w, y_l; \theta)\\
    = & -a_w \nabla_{\theta}\log\pi_{\theta}(y_w|x) + a_l\nabla_{\theta}\log\pi_{\theta}(y_l|x),
\end{split}
\end{equation}
where  the asymmetric weights are $a_w = \lambda_w\sigma(c_w)\sigma(1-c_w)$, and $a_l = \lambda_l\sigma(c_l)\sigma(1-c_l)$, with $c_w = \beta\left( \log\frac{\pi_{\theta}(y_w|x)}{\pi_{\text{ref}}(y_w|x)}-z_0\right)$ and $c_l = \beta\left( \log\frac{\pi_{\theta}(y_l|x)}{\pi_{\text{ref}}(y_l|x)}-z_0\right)$. Compared with DPO, these asymmetric weights provide stable gradients, guiding the model in adjusting the probability distribution effectively.

Moreover, to further investigate how KTO promotes probability shifts towards favorable outcomes during training, following~\cite{pal2024smaug} we re-arrange the derivative of the KTO loss as:
\begin{equation}
\small
\begin{split}
    &\nabla_{\theta}\mathcal{L}_{KTO}(x, y_w, y_l; \theta)\\
    =& \sum_{k = 1}^K \nabla_{\theta}\left[ -a_w \log\pi_{\theta}(t_k|y_w^{<k}, x )+a_l\log\pi_{\theta}(t_k|y_l^{<k}, x ) \right].\\
\end{split}
\label{eqn: kto-derivative-token}
\end{equation}

With the above formula,  each term in the derivative expression Eqn.~\eqref{eqn: kto-derivative-token} becomes:
\begin{equation*}
\small
\begin{split}
    &\nabla_{g_j}\left[ -a_w \log\pi_{\theta}(t_k|y_w^{<k}, x )+a_l\log\pi_{\theta}(t_k|y_l^{<k}, x ) \right]\\
    =&\left\{ \begin{array}{ll} 
          -a_w+a_l + a_w s_j^{y_w^{<k}, x} - a_l s_j^{y_l^{<k}, x} & \text{ $t_k=V_j$};\\
          a_w s_j^{y_w^{<k}, x}  - a_l s_j^{y_l^{<k}, x} & \text{ $t_k\neq V_j$}.
    \end{array}\right.
\end{split}
\end{equation*}
This analysis reveals that, given that $a_w > a_l$, it is evident that when the k-th token in $y_w$ matches $V_j$, the gradient is negative, whereas, for other vocabulary elements ($t_k\neq V_j$), the gradient is positive. Consequently, minimizing the KTO loss encourages an increase in the logits of the correct token, thereby effectively addressing DPO's second issue as discussed in Section~\ref{sec:DPO-failure}.

%% file: subfiles/4_Experiment.tex
% !TEX root = ../main.tex

\section{Experiment}
\begin{table}[h]
\centering
\vspace{-0.3cm}
\caption{Dataset Statistics}
\vspace{-0.3cm}
\fontsize{9}{10}\selectfont
\begin{tabular}{@{}c|ccc@{}}
\toprule
Dataset      & \# Queries & \# Tools & Multi-tool?   \\ \midrule
API-Bank L-1 & 399           & 49          & $\usym{2717}$ \\
API-Bank L-2 & 127           & 28          & $\usym{2717}$ \\
Tool-Alpaca  & 114           & 41          & $\usym{2714}$ \\
Seal-Tools   & 294           & 1084        & $\usym{2717}$ \\
Nexus Raven  & 318        & 65          & $\usym{2717}$ \\ \bottomrule
\end{tabular}
\label{tb:statistic}
\vspace{-0.3cm}
\end{table}

\begin{table*}[]
\centering
\vspace{-0.2cm}
\caption{Tool Calling Performance with HiTEC-ICL, with the best performance marked in \textbf{bold} and the proposed approach highlighted in \colorbox[HTML]{ECF4FF}{blue}.}
\vspace{-0.2cm}
\fontsize{9}{10}\selectfont
\setlength{\tabcolsep}{2pt} % 调小列间距
\begin{tabular}{@{}cccccccccccc|cc@{}}
\toprule
\multicolumn{12}{c|}{Dataset (F1 Name $|$ F1 Name +  Parameter)}                                                                                                                                                                                                                                                                                                                                                                                                                                     & \multicolumn{2}{c}{F1 Average}                                                                  \\ \midrule
Model                         & Method                              & \multicolumn{2}{c}{\begin{tabular}[c]{@{}c@{}}API-Bank\\ L-1\end{tabular}}      & \multicolumn{2}{c}{\begin{tabular}[c]{@{}c@{}}API-Bank\\ L-2\end{tabular}}      & \multicolumn{2}{c}{Tool-Alpaca}                                                 & \multicolumn{2}{c}{\begin{tabular}[c]{@{}c@{}}Seal-Tools\\ (Single-Tool)\end{tabular}} & \multicolumn{2}{c|}{\begin{tabular}[c]{@{}c@{}}Nexus\\ Raven\end{tabular}}      & Name                                   & \begin{tabular}[c]{@{}c@{}}Name+\\ Param.\end{tabular} \\ \midrule
                              & Vanilla                             & 94.77                                  & 86.66                                  & 71.09                                  & 70.43                                  & 87.12                                  & 60.75                                  & 94.86                                      & 88.55                                     & \textbf{94.13}                         & 82.92                                  & 88.39                                  & 77.86                                                  \\
                              & + CoT                               & 91.70                                  & 84.84                                  & 67.86                                  & 65.74                                  & 80.95                                  & 56.46                                  & 91.50                                      & 86.07                                     & 92.18                                  & \textbf{85.00}                         & 84.84                                  & 75.62                                                  \\
                              & + Function Calling                  & 84.03                                  & 81.40                                  & \textbf{78.49}                         & 62.32                                  & 87.11                                  & \textbf{62.16}                         & 84.94                                      & 84.88                                     & \textbackslash{}                       & \textbackslash{}                       & \textbackslash{}                       & \textbackslash{}                                       \\
\multirow{-4}{*}{GPT-4-Turbo} & \cellcolor[HTML]{ECF4FF}+ HiTEC-ICL & \cellcolor[HTML]{ECF4FF}\textbf{94.96} & \cellcolor[HTML]{ECF4FF}\textbf{88.44} & \cellcolor[HTML]{ECF4FF}71.21          & \cellcolor[HTML]{ECF4FF}\textbf{72.68} & \cellcolor[HTML]{ECF4FF}\textbf{87.35} & \cellcolor[HTML]{ECF4FF}61.95          & \cellcolor[HTML]{ECF4FF}\textbf{96.00}     & \cellcolor[HTML]{ECF4FF}\textbf{89.37}    & \cellcolor[HTML]{ECF4FF}92.81          & \cellcolor[HTML]{ECF4FF}83.15          & \cellcolor[HTML]{ECF4FF}\textbf{88.47} & \cellcolor[HTML]{ECF4FF}\textbf{79.12}                 \\ \midrule
                              & Vanilla                             & 69.19                                  & 67.73                                  & 75.54                                  & 48.91                                  & 40.46                                  & 24.86                                  & 86.54                                      & 82.51                                     & 14.83                                  & 17.44                                  & 57.31                                  & 48.29                                                  \\
                              & + CoT                               & 43.21                                  & 45.05                                  & 52.31                                  & 45.65                                  & 60.58                                  & 42.07                                  & 76.89                                      & 75.16                                     & 64.44                                  & 52.22                                  & 59.49                                  & 52.03                                                  \\
\multirow{-3}{*}{Llama3-8B}   & \cellcolor[HTML]{ECF4FF}+ HiTEC-ICL & \cellcolor[HTML]{ECF4FF}\textbf{75.93} & \cellcolor[HTML]{ECF4FF}\textbf{70.55} & \cellcolor[HTML]{ECF4FF}\textbf{77.30} & \cellcolor[HTML]{ECF4FF}\textbf{56.01} & \cellcolor[HTML]{ECF4FF}\textbf{85.92} & \cellcolor[HTML]{ECF4FF}\textbf{55.49} & \cellcolor[HTML]{ECF4FF}\textbf{94.40}     & \cellcolor[HTML]{ECF4FF}\textbf{87.27}    & \cellcolor[HTML]{ECF4FF}\textbf{82.07} & \cellcolor[HTML]{ECF4FF}\textbf{58.30} & \cellcolor[HTML]{ECF4FF}\textbf{83.12} & \cellcolor[HTML]{ECF4FF}\textbf{65.52}                 \\ \midrule
                              & Vanilla                             & 73.06                                  & 63.62                                  & 78.23                                  & 50.25                                  & 84.06                                  & 50.79                                  & 94.97                                      & 86.53                                     & \textbf{87.93}                         & 72.17                                  & 83.65                                  & 64.67                                                  \\
                              & + CoT                               & 65.22                                  & 61.61                                  & 75.54                                  & 48.52                                  & 63.31                                  & 42.57                                  & 88.55                                      & 80.36                                     & 76.71                                  & 67.32                                  & 73.87                                  & 60.08                                                  \\
\multirow{-3}{*}{Llama3.1-8B} & \cellcolor[HTML]{ECF4FF}+ HiTEC-ICL & \cellcolor[HTML]{ECF4FF}\textbf{75.07} & \cellcolor[HTML]{ECF4FF}\textbf{64.61} & \cellcolor[HTML]{ECF4FF}\textbf{82.23} & \cellcolor[HTML]{ECF4FF}\textbf{51.29} & \cellcolor[HTML]{ECF4FF}\textbf{85.92} & \cellcolor[HTML]{ECF4FF}\textbf{55.94} & \cellcolor[HTML]{ECF4FF}\textbf{96.95}     & \cellcolor[HTML]{ECF4FF}\textbf{87.94}    & \cellcolor[HTML]{ECF4FF}87.76          & \cellcolor[HTML]{ECF4FF}\textbf{72.35} & \cellcolor[HTML]{ECF4FF}\textbf{85.59} & \cellcolor[HTML]{ECF4FF}\textbf{66.42}                 \\ \bottomrule
\end{tabular}
\label{tb:main_icl}
\vspace{-0.3cm}
\end{table*}

\subsection{Experiment Setup}
We here describe the datasets, metrics, and baselines used in experiments.

\textbf{Datasets.} We conduct an evaluation on several benchmark tool calling datasets: API-Bank \citep{li2023api}, Tool Alpaca \citep{tang2023toolalpaca}, Seal-Tools \citep{wu2024seal}, and Nexus Raven \citep{srinivasan2023nexusraven}. All datasets are preprocessed in the same way as Lin et al. \cite{lin2024hammer}. The statistics of the processed datasets are summarized in Table \ref{tb:statistic}. 

\textbf{Metrics.} We evaluate two performance metrics in the form of F1 score: correctness of tool names (F1 Name) and correctness of tool names + tool parameters (F1 Name + Parameter). 

\textbf{Baselines.} We use GPT-3.5-Turbo-0125, GPT-4-Turbo (GPT series), Llama3-8B, Llama3-70B (Llama3 series), Llama3.1-8B, Llama3.1-70 (Llama3.1-series) models as base models to study the effect of HiTEC-ICL. The vanilla version of the base models, base models deployed with zero-shot CoT \cite{wei2022chain}, and native tool calling integrated base models \footnote{\url{https://platform.openai.com/docs/guides/function-calling}} (only GPT-series) are used as baselines. For HiTEC-KTO, we use Llama3-8B,  Llama3-70B (Llama3 series), Llama3.1-8B, Llama3.1-70 (Llama3.1-series), Qwen2.5-0.5B, Qwen2.5-1.5B, Qwen2.5-3B and Qwen2.5-7B (Qwen2.5 series) as base models, and consider the vanilla base models of Llama series and Hammer2.0 release \footnote{\url{https://huggingface.co/collections/MadeAgents/hammer20-66f4dee539f7b2c95224012a}} (which is tuned on Qwen2.5 series \cite{lin2024hammer}) as baselines. The implementation details are in Appendix \ref{app:imp_details}.

\subsection{Main Results}

\subsubsection{Effectiveness of HiTEC-ICL}

We now evaluate HiTEC's efficacy across datasets and base models in ICL setting. The method consistently enhances tool-calling performance as shown in Table \ref{tb:main_icl}. Experiments under more settings are deferred to Appendix \ref{app:exp} due to limited space.

HiTEC-ICL demonstrates robust performance improvements across diverse tool-learning benchmarks. Model capability critically influences HiTEC-ICL’s efficacy. Smaller models (Llama3-8B) achieve the most pronounced gains, with Name + Parameter improving by over 30 points on Nexus Raven (58.30 vs. 17.44), indicating that explicit error guidance compensates for limited reasoning capacity. Larger models (GPT-4-Turbo, Llama3.1-70B) exhibit subtler but consistent improvements, leveraging checklists to refine already strong baseline performance. Notably, HiTEC-ICL outperforms CoT and vanilla tool calling in most cases, with the largest margins on Seal-Tools validating its structured error mitigation approach.

\subsubsection{Effectiveness of HiTEC-KTO}

\begin{table*}[]
\vspace{-0.3cm}
\caption{Tool Calling Performance with HiTEC-KTO, with the best performance marked in \textbf{bold} and the proposed approach highlighted in \colorbox[HTML]{ECF4FF}{blue}}
%\vspace{-0.2cm}
\fontsize{9}{10}\selectfont
\setlength{\tabcolsep}{2pt} % 调小列间距
\begin{tabular}{@{}ccccccccccccc|cc@{}}
\toprule
\multicolumn{13}{c|}{Dataset (F1 Name $|$ F1 Name + Parameter)}                                                                                                                                                                                                                                                                                                                                                                                                                                                                                                                                                                                                           & \multicolumn{2}{c}{F1 Average}                                                                   \\ \midrule
\multicolumn{1}{c|}{\begin{tabular}[c]{@{}c@{}}Model\\ Series\end{tabular}}                      & Method                                                                                         & Model                                & \multicolumn{2}{c}{\begin{tabular}[c]{@{}c@{}}API-Bank\\ L-1\end{tabular}}      & \multicolumn{2}{c}{\begin{tabular}[c]{@{}c@{}}API-Bank\\ L-2\end{tabular}}      & \multicolumn{2}{c}{Tool-Alpaca}                                                 & \multicolumn{2}{c}{\begin{tabular}[c]{@{}c@{}}Seal-Tools\\ (Single-Tool)\end{tabular}} & \multicolumn{2}{c|}{\begin{tabular}[c]{@{}c@{}}Nexus\\ Raven\end{tabular}}      & Name                                   & \begin{tabular}[c]{@{}c@{}}Name +\\ Param.\end{tabular} \\ \midrule
\multicolumn{1}{c|}{}                                                                            &                                                                                                & Llama3.1-8B                          & 73.06                                  & 63.62                                  & 78.23                                  & 50.25                                  & 84.06                                  & 50.79                                  & 94.97                                      & \textbf{86.53}                            & 87.93                                  & 72.17                                  & 83.65                                  & 64.67                                                   \\
\multicolumn{1}{c|}{}                                                                            & \multirow{-2}{*}{Baseline}                                                                     & Llama3.1-70B                         & 90.15                                  & 76.42                                  & 80.34                                  & 62.20                                  & 86.03                                  & 55.83                                  & \textbf{97.42}                             & 86.19                                     & 93.75                                  & 82.72                                  & 89.54                                  & 72.67                                                   \\ \cmidrule(l){2-15} 
\multicolumn{1}{c|}{}                                                                            & \cellcolor[HTML]{ECF4FF}                                                                       & \cellcolor[HTML]{ECF4FF}Llama3.1-8B  & \cellcolor[HTML]{ECF4FF}\textbf{87.47} & \cellcolor[HTML]{ECF4FF}\textbf{80.99} & \cellcolor[HTML]{ECF4FF}\textbf{85.61} & \cellcolor[HTML]{ECF4FF}\textbf{61.02} & \cellcolor[HTML]{ECF4FF}\textbf{84.18} & \cellcolor[HTML]{ECF4FF}\textbf{56.45} & \cellcolor[HTML]{ECF4FF}\textbf{94.44}     & \cellcolor[HTML]{ECF4FF}86.18             & \cellcolor[HTML]{ECF4FF}\textbf{89.98} & \cellcolor[HTML]{ECF4FF}\textbf{82.28} & \cellcolor[HTML]{ECF4FF}\textbf{88.34} & \cellcolor[HTML]{ECF4FF}\textbf{73.38}                  \\
\multicolumn{1}{c|}{\multirow{-4}{*}{\begin{tabular}[c]{@{}c@{}}Llama3.1\\ Series\end{tabular}}} & \multirow{-2}{*}{\cellcolor[HTML]{ECF4FF}\begin{tabular}[c]{@{}c@{}}HiTEC-\\ KTO\end{tabular}} & \cellcolor[HTML]{ECF4FF}Llama3-70B   & \cellcolor[HTML]{ECF4FF}\textbf{90.78} & \cellcolor[HTML]{ECF4FF}\textbf{77.07} & \cellcolor[HTML]{ECF4FF}\textbf{86.44} & \cellcolor[HTML]{ECF4FF}\textbf{65.14} & \cellcolor[HTML]{ECF4FF}\textbf{86.67} & \cellcolor[HTML]{ECF4FF}\textbf{57.32} & \cellcolor[HTML]{ECF4FF}\textbf{98.14}     & \cellcolor[HTML]{ECF4FF}\textbf{90.01}    & \cellcolor[HTML]{ECF4FF}\textbf{94.84} & \cellcolor[HTML]{ECF4FF}\textbf{82.87} & \cellcolor[HTML]{ECF4FF}\textbf{91.37} & \cellcolor[HTML]{ECF4FF}\textbf{74.48}                  \\ \midrule
\multicolumn{1}{c|}{}                                                                            &                                                                                                & Hammer2-0.5B                         & 71.20                                  & 59.03                                  & 43.32                                  & 38.22                                  & 64.46                                  & \textbf{41.83}                         & 93.86                                      & 83.03                                     & 64.72                                  & 45.52                                  & 67.51                                  & 53.53                                                   \\
\multicolumn{1}{c|}{}                                                                            &                                                                                                & Hammer2-1.5B                         & 88.63                                  & 79.26                                  & 80.51                                  & \textbf{62.82}                         & 80.74                                  & 51.88                                  & 96.10                                      & 87.16                                     & 85.85                                  & 63.76                                  & 86.37                                  & 68.98                                                   \\
\multicolumn{1}{c|}{}                                                                            &                                                                                                & Hammer2-3B                           & 88.63                                  & 79.04                                  & 77.11                                  & \textbf{57.58}                         & 78.23                                  & 53.28                                  & 93.08                                      & 85.60                                     & \textbf{89.14}                         & \textbf{66.71}                         & 85.24                                  & 68.44                                                   \\
\multicolumn{1}{c|}{}                                                                            & \multirow{-4}{*}{Baseline}                                                                     & Hammer2-7B                           & 88.91                                  & 81.28                                  & 75.96                                  & 58.36                                  & 81.74                                  & 57.07                                  & 94.62                                      & 87.84                                     & 90.76                                  & \textbf{80.96}                         & 86.40                                  & 73.10                                                   \\ \cmidrule(l){2-15} 
\multicolumn{1}{c|}{}                                                                            & \cellcolor[HTML]{ECF4FF}                                                                       & \cellcolor[HTML]{ECF4FF}Qwen2.5-0.5B & \cellcolor[HTML]{ECF4FF}\textbf{88.29} & \cellcolor[HTML]{ECF4FF}\textbf{78.34} & \cellcolor[HTML]{ECF4FF}\textbf{81.76} & \cellcolor[HTML]{ECF4FF}\textbf{54.52} & \cellcolor[HTML]{ECF4FF}\textbf{73.54} & \cellcolor[HTML]{ECF4FF}38.76          & \cellcolor[HTML]{ECF4FF}\textbf{96.80}     & \cellcolor[HTML]{ECF4FF}\textbf{88.18}    & \cellcolor[HTML]{ECF4FF}\textbf{82.83} & \cellcolor[HTML]{ECF4FF}\textbf{63.89} & \cellcolor[HTML]{ECF4FF}\textbf{84.64} & \cellcolor[HTML]{ECF4FF}\textbf{64.74}                  \\
\multicolumn{1}{c|}{}                                                                            & \cellcolor[HTML]{ECF4FF}                                                                       & \cellcolor[HTML]{ECF4FF}Qwen2.5-1.5B & \cellcolor[HTML]{ECF4FF}\textbf{88.92} & \cellcolor[HTML]{ECF4FF}\textbf{79.46} & \cellcolor[HTML]{ECF4FF}\textbf{81.36} & \cellcolor[HTML]{ECF4FF}60.66          & \cellcolor[HTML]{ECF4FF}\textbf{82.27} & \cellcolor[HTML]{ECF4FF}\textbf{52.26} & \cellcolor[HTML]{ECF4FF}\textbf{96.99}     & \cellcolor[HTML]{ECF4FF}\textbf{89.75}    & \cellcolor[HTML]{ECF4FF}\textbf{86.37} & \cellcolor[HTML]{ECF4FF}\textbf{64.40} & \cellcolor[HTML]{ECF4FF}\textbf{87.18} & \cellcolor[HTML]{ECF4FF}\textbf{69.31}                  \\
\multicolumn{1}{c|}{}                                                                            & \cellcolor[HTML]{ECF4FF}                                                                       & \cellcolor[HTML]{ECF4FF}Qwen2.5-3B   & \cellcolor[HTML]{ECF4FF}\textbf{89.07} & \cellcolor[HTML]{ECF4FF}\textbf{80.16} & \cellcolor[HTML]{ECF4FF}\textbf{85.51} & \cellcolor[HTML]{ECF4FF}54.61          & \cellcolor[HTML]{ECF4FF}\textbf{85.42} & \cellcolor[HTML]{ECF4FF}\textbf{56.08} & \cellcolor[HTML]{ECF4FF}\textbf{96.28}     & \cellcolor[HTML]{ECF4FF}\textbf{89.51}    & \cellcolor[HTML]{ECF4FF}83.05          & \cellcolor[HTML]{ECF4FF}62.35          & \cellcolor[HTML]{ECF4FF}\textbf{87.87} & \cellcolor[HTML]{ECF4FF}\textbf{68.54}                  \\
\multicolumn{1}{c|}{\multirow{-8}{*}{\begin{tabular}[c]{@{}c@{}}Qwen2.5\\ Series\end{tabular}}}  & \multirow{-4}{*}{\cellcolor[HTML]{ECF4FF}\begin{tabular}[c]{@{}c@{}}HiTEC-\\ KTO\end{tabular}} & \cellcolor[HTML]{ECF4FF}Qwen2.5-7B   & \cellcolor[HTML]{ECF4FF}\textbf{89.38} & \cellcolor[HTML]{ECF4FF}\textbf{81.67} & \cellcolor[HTML]{ECF4FF}\textbf{88.00} & \cellcolor[HTML]{ECF4FF}\textbf{59.93} & \cellcolor[HTML]{ECF4FF}\textbf{87.63} & \cellcolor[HTML]{ECF4FF}\textbf{59.19} & \cellcolor[HTML]{ECF4FF}\textbf{96.27}     & \cellcolor[HTML]{ECF4FF}\textbf{89.29}    & \cellcolor[HTML]{ECF4FF}\textbf{91.70} & \cellcolor[HTML]{ECF4FF}78.17          & \cellcolor[HTML]{ECF4FF}\textbf{90.60} & \cellcolor[HTML]{ECF4FF}\textbf{73.65}                  \\ \bottomrule
\end{tabular}
\label{tb:main_kto}
\vspace{-0.2cm}
\end{table*}

\begin{table}[]
\centering
%\vspace{-0.3cm}
\caption{Ablation Study on the Hierarchical Error Checklist of HiTEC-KTO}
%\vspace{-0.3cm}
\fontsize{8}{9}\selectfont
\setlength{\tabcolsep}{1.5pt} % 调小列间距
\begin{tabular}{@{}cccccccc@{}}
\toprule
\multicolumn{8}{c}{Base Model (F1 Name | F1 Name + Parameter)}                                                                                                                                                                           \\ \midrule
Dataset                      & Method         & \multicolumn{2}{c}{\begin{tabular}[c]{@{}c@{}}Qwen2.5\\ 1.5B\end{tabular}} & \multicolumn{2}{c|}{\begin{tabular}[c]{@{}c@{}}Qwen2.5\\ 3B\end{tabular}} & \multicolumn{2}{c}{Average}     \\ \midrule
\multirow{3}{*}{Tool Alpaca} & HiTEC-KTO      & \textbf{82.27}                       & \textbf{52.26}                      & \textbf{85.42}            & \multicolumn{1}{c|}{\textbf{56.08}}           & \textbf{83.30} & \textbf{53.30} \\
                             & w/o Local EC   & 60.61                                & 28.97                               & 77.62                     & \multicolumn{1}{c|}{51.61}                    & 69.71          & 41.08          \\
                             & w/o Glb-loc EC & 22.83                                & 16.03                               & 73.45                     & \multicolumn{1}{c|}{47.75}                    & 46.38          & 31.40          \\ \midrule
\multirow{3}{*}{Seal-Tools}  & HiTEC-KTO      & \textbf{96.99}                       & \textbf{89.75}                      & \textbf{96.28}            & \multicolumn{1}{c|}{\textbf{89.51}}           & \textbf{96.62} & \textbf{89.48} \\
                             & w/o Local EC   & 87.75                                & 68.99                               & 92.75                     & \multicolumn{1}{c|}{82.60}                    & 90.31          & 76.85          \\
                             & w/o Glb-loc EC & 41.86                                & 37.34                               & 83.98                     & \multicolumn{1}{c|}{77.99}                    & 60.01          & 55.29          \\ \bottomrule
\end{tabular}
\label{tb:ablation_kto}
%\vspace{-0.3cm}
\end{table}

\textbf{KTO Training Data.} We generate negative samples based on the xlam-function-calling-60k dataset \cite{liuapigen}. One incorrect tool calling answer is generated for each of the queries in the xlam-function-calling-60k dataset. We label the incorrect answer as "False" and the original groundtruth as "True" and perform KTO on the 12,000 samples. Table \ref{tb:main_kto} demonstrates the experiment results over various of open-sourced models. Results under more settings are deferred to Appendix \ref{app:exp} due to limited space. 

It can be found that HiTEC-KTO significantly enhances tool-calling capabilities across open-source models by leveraging error checklists to generate targeted negative examples for preference-based fine-tuning. Crucially, HiTEC-KTO enables smaller models to rival larger counterparts—the 1.5B Qwen2.5 model surpasses the 3B baseline Hammer in many cases after being fine-tuned with HiTECH-KTO. Performance trends also correlate with dataset complexity. On multi-tool calling dataset Tool Alpaca, HiTEC-KTO boosts parameter accuracy by 6-25 points on F1 across 8B Llama series, demonstrating its efficacy in resolving tool-specific ambiguities. It also provides 8B Llama series with a 4-10 points performance increase on the Nexus Raven dataset, which has complex and long queries. 

These results underscore HiTEC-KTO’s ability to democratize advanced tool-learning capabilities across model sizes and dataset complexity, balancing error avoidance with functional precision through structured checklist-driven optimization.

\subsection{Ablation Study}
To assess the impact of HiTEC’s hierarchical design, we conducted an ablation study comparing configurations that utilize the Global Error Checklist (Glb EC) versus the combined Global-Local Error Checklist (Glblc EC) across two datasets—Tool Alpaca and Seal-Tools—as well as multiple base models. The results are presented in Table \ref{tb:ablation_kto} and Appendix \ref{app:exp} due to limited space. For HiTEC-KTO variants, w/o Local EC indicates that KTO was performed on negative samples generated solely with the global error checklist, omitting the local error checklist. w/o Glblc EC refers to the baseline model without hierarchical error correction.

HiTEC-KTO significantly outperforms w/o Glblc EC, particularly in smaller models. Notably, HiTEC Qwen2.5-1.5B achieves performance comparable to its vanilla counterpart (w/o Glb-loc EC). Removing the local error checklist has a substantial negative impact on HiTEC-KTO, as global errors alone capture only general mistakes, whereas generating high-quality negative examples requires tool-specific error instructions.

\begin{figure}[t]
%\vspace{-0.35cm}
\centering
\includegraphics[width=0.95\linewidth]{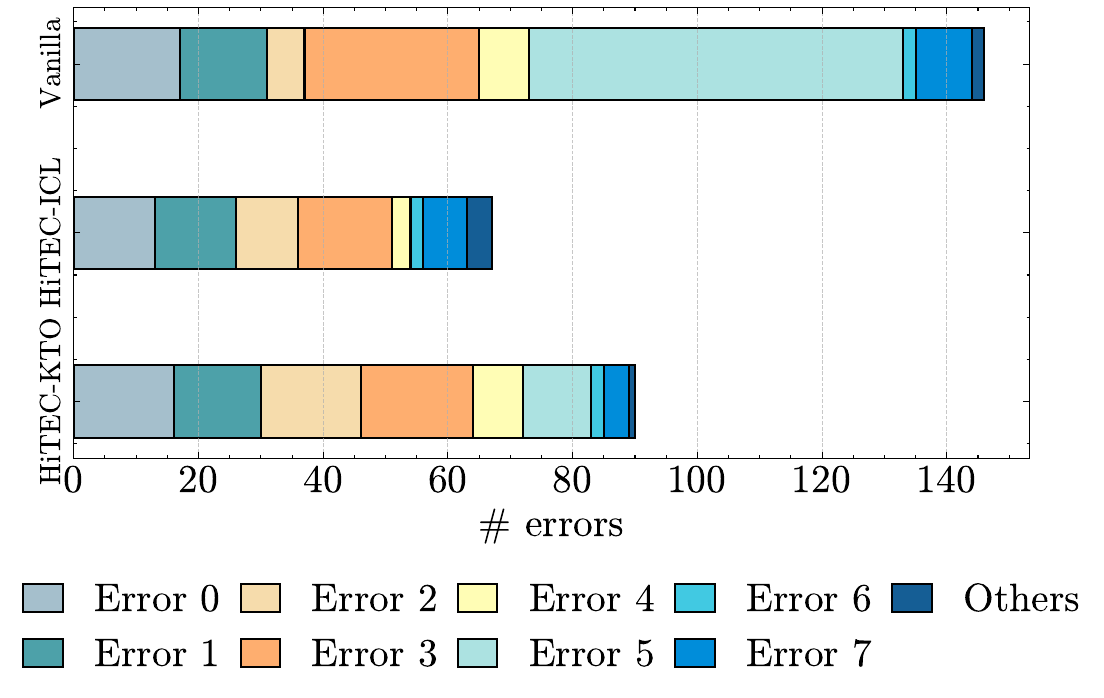}
\vspace{-0.3cm}
\caption{Analysis on Error Distribution}
\label{fig:error_distribution}
\vspace{-0.45cm}
\end{figure}

As shown in the table, incorporating the local error checklist generally improves accuracy for both tool name identification and parameter filling compared to using only the global checklist. This improvement is primarily attributed to more precise parameter filling, as evidenced by the greater increase in the F1 Name + Parameter metric (second column) compared to F1 Name (first column). For instance, on Tool Alpaca with Qwen2.5-1.5B, F1 Name + Parameter relatively improves by 86\%, whereas the relative improvement of F1 Name is only 37\%. This result highlights the effectiveness of hierarchical checklists in addressing tool-specific local errors. 

In conclusion, HiTEC’s hierarchical structure effectively mitigates tool-specific errors, particularly enhancing parameter accuracy in smaller models.

\begin{figure*}[t]
\vspace{-0.45cm}
\centering
\includegraphics[width=0.95\linewidth]{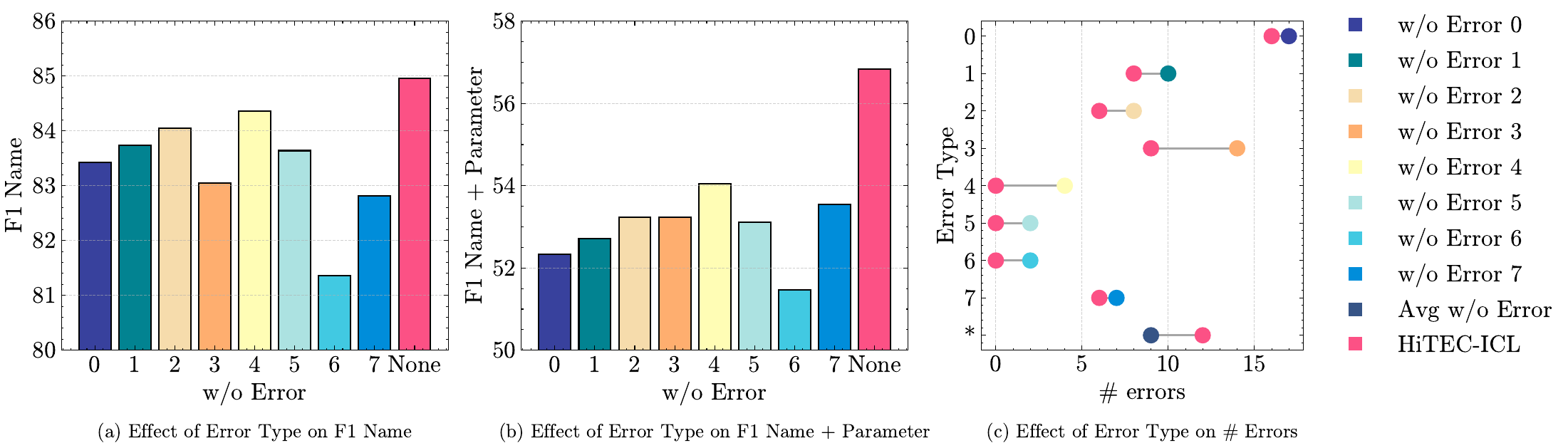}
\vspace{-0.25cm}
\caption{Effect of Error Types}
\vspace{-0.3cm}
\end{figure*}

\subsection{Look Into the Errors}
In this subsection, we further look into errors and conduct error-level evaluations. 
\subsubsection{Analysis on Error Distribution}

We analyze error type distributions across the vanilla base model, HiTEC-ICL, and HiTEC-KTO configurations on Tool Alpaca (Llama3-8B) to quantify the framework’s impact. Since multiple errors can co-exist in a single response, we count all occurrences. The results are shown in \ref{fig:error_distribution}.

In vanilla setting, parameter-level errors (Error 1–4) and format errors (Error 5) account for the majority of failures. Formatting errors (Error 5), in particular, are prevalent, reflecting baseline weaknesses in syntactical consistency. However, with HiTEC, the percentage of formatting errors drops significantly, demonstrating the global checklist’s effectiveness in enforcing output structure.

Parameter-filling errors (Error 1–4) also undergo a notable redistribution. In the vanilla configuration, Empty Parameter Value Error (Error 3) is the most frequent, but under HiTEC, the distribution becomes more balanced, with no single error type dominating. This indicates that the hierarchical error checks address a broader range of parameter-related issues, reducing reliance on any single mitigation strategy.

\subsubsection{Effect of Error Types}
\label{sec:error_type}
To evaluate the contribution of individual error types in the HiTEC framework, we conduct an experiment in which each error type is iteratively excluded from the HiTEC, and the performance of HiTEC-ICL is assessed. The results for HiTEC-ICL-Llama3.1-70B on the Tool Alpaca dataset are presented in Figure \ref{fig:error_type}, while additional results under different settings can be found in Appendix \ref{app:exp} due to limited space.

As illustrated in Figure \ref{fig:error_type} (a-b), the exclusion of tool-level errors (e.g., Errors 0, 6, and 7) significantly reduces performance, highlighting their critical role in the multi-tool selection feature of the Tool Alpaca dataset. In contrast, parameter-centric errors primarily affect the Seal-Tools dataset (see Appendix \ref{app:exp}). Notably, the removal of Errors 3 and 4 results in a substantial performance decline across multiple base models. Furthermore, Error 5 is found to be universally essential, as its presence ensures the generation of parsable outputs, which are crucial for accurate tool calling.

Additionally, we examine changes in the number of errors (\# errors) when specific error types are removed. Figure \ref{fig:error_type} (c) demonstrates that the number of errors in categories 0-7 increases when the corresponding error type is omitted from the checklist. This observation confirms that a defined error in HiTEC can help reduce such a type of error. The symbol * denotes other types of errors, while "Avg w/o Error" represents the average number of errors within the "Others" category of the w/o variants. HiTEC-ICL exhibits a slight increase in errors within the "Others" category compared to its variants, which is a reasonable outcome given that the removed errors may contribute to new, previously unclassified errors.

\subsection{Cost Analysis}

In thie section, we further conduct an evaluation of HiTEC-ICL's inference overhead. Specifically, we measure the number of prompt tokens and generated tokens on the Tool-Alpaca dataset, comparing HiTEC-ICL against the vanilla model, CoT, and HiTEC-ICL with only a global checklist. Additionally, we assessed the cost per query in US dollars for GPT-based models and execution time for Llama-based models. All experiments were conducted on four NVIDIA A100 GPUs. The results are shown in Table \ref{tb:cost_close} and Table \ref{tb:cost_open}.

The results indicate that while HiTEC-ICL incurs the highest cost, the absolute expense remains within a reasonable range. Notably, the substantial performance gains observed on Llama-based models justify the increase in computational cost. Moreover, HiTEC-ICL (w/o local) strikes a strong balance between efficiency and effectiveness, offering lower costs while maintaining superior performance compared to the baselines.

\begin{table}[]
\centering
\vspace{-0.3cm}
\caption{Cost Analysis of HiTEC-ICL on Closed-sourced Models}
\fontsize{8.5}{9}\selectfont
\setlength{\tabcolsep}{2.5pt} % 调小列间距
\begin{tabular}{@{}cccccc@{}}
\toprule
Model                                                                             & Method                                                           & \begin{tabular}[c]{@{}c@{}}Prompt\\ Tokens\end{tabular} & \begin{tabular}[c]{@{}c@{}}Generated\\ Tokens\end{tabular} & \begin{tabular}[c]{@{}c@{}}Cost per\\ Case (\$)\end{tabular} & \begin{tabular}[c]{@{}c@{}}F1 Name\\  + Param.\end{tabular} \\ \midrule
\multirow{4}{*}{\begin{tabular}[c]{@{}c@{}}GPT-3.5-\\ Turbo-\\ 0125\end{tabular}} & Vanilla                                                          & 77,374                                                  & 3,759                                                      & 0.0004                                                       & 56.47                                                       \\
                                                                                  & CoT                                                              & 82,048                                                  & 15,401                                                     & 0.0006                                                       & 58.43                                                       \\
                                                                                  & \begin{tabular}[c]{@{}c@{}}HiTEC-ICL\\  (w/o local)\end{tabular} & \textbf{91,510}                                         & \textbf{3,747}                                             & \textbf{0.0005}                                              & \textbf{57.98}                                              \\
                                                                                  & HiTEC-ICL                                                        & \textbf{320,830}                                        & \textbf{7,128}                                             & \textbf{0.0015}                                              & \textbf{58.79}                                              \\ \midrule
\multirow{4}{*}{\begin{tabular}[c]{@{}c@{}}GPT-4-\\ Turbo\end{tabular}}           & Vanilla                                                          & 77,374                                                  & 3,678                                                      & 0.0078                                                       & 60.75                                                       \\
                                                                                  & CoT                                                              & 81,706                                                  & 25,429                                                     & 0.0127                                                       & 56.46                                                       \\
                                                                                  & \begin{tabular}[c]{@{}c@{}}HiTEC-ICL\\  (w/o local)\end{tabular} & \textbf{103,252}                                        & \textbf{3,680}                                             & \textbf{0.0100}                                              & \textbf{60.67}                                              \\
                                                                                  & HiTEC-ICL                                                        & \textbf{321,816}                                        & \textbf{7,344}                                             & \textbf{0.0302}                                              & \textbf{61.95}                                              \\ \bottomrule
\end{tabular}
\label{tb:cost_close}
\vspace{-0.2cm}
\end{table}

\begin{table}[]
\centering
\caption{Cost Analysis of HiTEC-ICL on Open-sourced Models}
\fontsize{8.5}{8}\selectfont
\setlength{\tabcolsep}{2.5pt} % 调小列间距
\begin{tabular}{@{}cccccc@{}}
\toprule
Model                                                                   & Method                                                          & \begin{tabular}[c]{@{}c@{}}Prompt\\ Tokens\end{tabular} & \begin{tabular}[c]{@{}c@{}}Generated\\ Tokens\end{tabular} & \begin{tabular}[c]{@{}c@{}}Cost per\\ Case\\ (second)\end{tabular} & \begin{tabular}[c]{@{}c@{}}F1 Name \\ + Param.\end{tabular} \\ \midrule
\multirow{4}{*}{\begin{tabular}[c]{@{}c@{}}Llama3-\\ 8B\end{tabular}}   & Vanilla                                                         & 81,260                                                  & 3,786                                                      & 0.0360                                                             & 24.86                                                       \\
                                                                        & CoT                                                             & 102,997                                                 & 21,191                                                     & 0.0701                                                             & 42.07                                                       \\
                                                                        & \begin{tabular}[c]{@{}c@{}}HiTEC-ICL\\ (w/o local)\end{tabular} & \textbf{107,005}                                        & \textbf{3,653}                                             & \textbf{0.0454}                                                    & \textbf{52.39}                                              \\
                                                                        & HiTEC-ICL                                                       & \textbf{323,392}                                        & \textbf{7,532}                                             & \textbf{0.1320}                                                    & \textbf{55.49}                                              \\ \midrule
\multirow{4}{*}{\begin{tabular}[c]{@{}c@{}}Llama3.1-\\ 8B\end{tabular}} & Vanilla                                                         & 81,670                                                  & 4,196                                                      & 0.0368                                                             & 50.79                                                       \\
                                                                        & CoT                                                             & 108,503                                                 & 26,355                                                     & 0.0799                                                             & 42.57                                                       \\
                                                                        & \begin{tabular}[c]{@{}c@{}}HiTEC-ICL\\ (w/o local)\end{tabular} & \textbf{107,534}                                        & \textbf{4,182}                                             & \textbf{0.0464}                                                    & \textbf{52.33}                                              \\
                                                                        & HiTEC-ICL                                                       & \textbf{331,671}                                        & \textbf{8,800}                                             & \textbf{0.1370}                                                    & \textbf{55.94}                                              \\ \bottomrule
\end{tabular}
\label{tb:cost_open}
\vspace{-0.45cm}
\end{table}

\subsection{Study on Multi-turn/step Tool Calling}
To further evaluate HiTEC’s capability in handling interactive tool use, we design experiments to simulate multi-turn/step tool calling. In this setup, the responses from previous tool calls are treated as predefined contextual information for subsequent calls. We incorporate such intermediate responses into artificial conversations within our proposed framework, prompting the model to generate the next API call accordingly.

To ensure a comprehensive evaluation, we filtered multi-turn/step tool queries from the Seal-Tools dataset \cite{wu2024seal} and conducted experiments on HiTEC-ICL and HiTEC-KTO. The results, presented in Table \ref{tb:hitech-icl-multi} and \ref{tb:hitech-kto-multi}, demonstrate that while multi-turn tool use poses greater challenges, our approach consistently outperforms the baselines in most cases, highlighting its effectiveness in iterative tool interactions.

\begin{table}[]
\centering
\vspace{-0.2cm}
\caption{Tool Calling Performance with HiTEC-ICL on Seal-Tools Multi-turn/step}
\fontsize{9}{10}\selectfont
\setlength{\tabcolsep}{2.5pt} % 调小列间距
\begin{tabular}{@{}cccc@{}}
\toprule
Method                     & Model        & F1 Name        & \begin{tabular}[c]{@{}c@{}}F1 Name\\ + Param.\end{tabular} \\ \midrule
\multirow{3}{*}{Llama 3-8B}  & Vanilla    & 8.24               & 0.00              \\
                             & +CoT       & 8.42               & 5.40              \\
                             & +HiTEC-ICL & \textbf{43.01}     & \textbf{10.21}    \\ \midrule
\multirow{3}{*}{Llama3.1-8B} & Vanilla    & 47.51              & 10.77             \\
                             & +CoT       & 43.66              & 9.86              \\
                             & +HiTEC-ICL & \textbf{57.14}     & \textbf{15.87}    \\ \bottomrule
\end{tabular}
\label{tb:hitech-icl-multi}
\vspace{-0.1cm}
\end{table}

\begin{table}[]
\centering
\fontsize{9}{10}\selectfont
\setlength{\tabcolsep}{2.5pt} % 调小列间距
%\vspace{-0.3cm}
\caption{Tool Calling Performance with HiTEC-KTO on Seal-Tools Multi-turn/step}
\begin{tabular}{@{}cccc@{}}
\toprule
Method                     & Model        & F1 Name        & \begin{tabular}[c]{@{}c@{}}F1 Name\\ + Param.\end{tabular} \\ \midrule
\multirow{2}{*}{Baseline}  & Llama3.1-8B  & 47.51          & 10.77                                                      \\
                           & Llama3.1-70B & 51.23          & 16.76                                                      \\ \midrule
\multirow{2}{*}{HiTEC-KTO} & Llama3.1-8B  & \textbf{67.15} & \textbf{14.09}                                             \\
                           & Llama3.1-70B & \textbf{70.31} & \textbf{18.81}                                             \\ \midrule
\multirow{4}{*}{Baseline}  & Hammer2-0.5B & 31.78          & 4.65                                                       \\
                           & Hammer2-1.5B & \textbf{47.70}          & 9.41                                                       \\
                           & Hammer2-3B   & 32.09          & 5.55                                                       \\
                           & Hammer2-7B   & 43.03          & \textbf{17.24}                                                      \\ \midrule
\multirow{4}{*}{HiTEC-KTO} & Qwen2.5-0.5B & \textbf{35.95} & \textbf{8.35}                                              \\
                           & Qwen2.5-1.5B & 44.70 & \textbf{11.15}                                             \\
                           & Qwen2.5-3B   & \textbf{54.67} & \textbf{14.00}                                             \\
                           & Qwen2.5-7B   & \textbf{58.15} & 14.65                                             \\ \bottomrule
\end{tabular}
\label{tb:hitech-kto-multi}
\vspace{-0.45cm}
\end{table}

%% file: subfiles/6_Conclusion.tex
% !TEX root = ../main.tex

\section{Conclusion}
In this work, we introduced the Hierarchical Tool Error Checklist (HiTEC) framework to enhance tool calling in large language models by systematically identifying and addressing common and tool-specific errors. By integrating both global and local error checklists, HiTEC effectively mitigates parameter mis-filling and format inconsistencies. We proposed two complementary approaches—HiTEC-ICL and HiTEC-KTO—to incorporate error feedback into the tool-calling process. HiTEC-ICL leverages structured error guidance during prompt formulation, while HiTEC-KTO utilizes targeted negative examples to fine-tune open-source models. Extensive experiments on multiple benchmark datasets demonstrate that both methods significantly improve tool name identification and parameter accuracy compared to existing baselines, with particularly notable gains in smaller models.

%% file: subfiles/7_Appendix.tex
\appendix
\clearpage  % 强制新页，但不会引入空白页
\onecolumn
\section{Prompt}
\label{app:prompt}

\begin{tcolorbox}[colback=bluee!10, colframe=bluee!80, title=Local Error Checklist Generation Prompt]

Task:  \\
You are given information about a tool and an example template of an error checklist. Your task is to generate an error checklist for the tool in the same format as the template. More specifically, for each error, you should:   \\
- Provide a **perfect query**. The query should be self-contained and contain all the necessary information for a correct tool call. For example: "Can you verify the access to the database named 'customer\_data'?"  \\
- Provide the **corresponding answer** from the model that evokes the error.  \\
- Provide an **error message** that describes what went wrong.  \\
- Provide a **thought** explaining how the error should be corrected. \\ 

Note: You should strictly follow the format of the template.

------

**Error Checklist Template**

Tool Information  

name: 'name\_of\_the\_tool'\ \\
description: 'description\_of\_the\_tool'\\
parameters: \{"parameter\_name\_1": \{"type": "type\_1", "description": "description\_of\_the\_parameter"\},  "parameter\_2": \{"type": "type\_2",  "description": "description\_of\_the\_parameter"\}\} required parameters: ["parameter\_1"]  
(Include other relevant information about the tool if necessary.)

---

Error 2: Missing Required Parameter Error  

Query: "a\_query\_that\_calls\_the\_tool"  

Function Calling Output: [\{\{"name": "name\_of\_the\_tool","arguments": \{\{"parameter\_2":"parameter\_value"\}\}\}\}]  

Error Message: \{\{"error": "MissingRequiredParameter","message": "The 'parameter\_1' parameter is required."\}\}

Thought of Error: Parameter 'parameter\_1' is missing. Ensure all required parameters ('parameter\_1') are included in the function call.

---

Error 3: Invalid Parameter Type Error  

Query: "a\_query\_that\_calls\_the\_tool"  

Function Calling Output:  
[
  \{\{
    "name": "name\_of\_the\_tool",
    "arguments": \{\{
      "parameter\_1": "parameter\_value",
      "parameter\_2": "parameter\_value (but not of type\_2)"
    \}\}
  \}\}
]  

Error Message:  
\{\{
  "error": "InvalidParameterType",
  "message": "The 'parameter\_2' is not of 'type\_2'."
\}\}

Thought of Error:  
Parameter 'parameter\_2' should be of type 'type\_2', but an invalid type was provided. Ensure all parameters match their expected types.

---

Error 4: Empty Parameter Value Error  

Query: "a\_query\_that\_calls\_the\_tool"  

Function Calling Output:  
[
  \{\{
    "name": "name\_of\_the\_tool",
    "arguments": \{\{
      "parameter\_1": "parameter\_value",
      "parameter\_2": ""
    \}\}
  \}\}
]  

Error Message:  
\{\{
  "error": "EmptyParameterValue",
  "message": "The 'parameter\_2' parameter cannot be empty."
\}\}
\end{tcolorbox}
\begin{tcolorbox}[colback=bluee!10, colframe=bluee!80, title=Local Error Checklist Generation Prompt (cont'd)]
Thought of Error:  
Parameter 'parameter\_2' has an empty value. It should not be empty as specified by the tool's requirements.

---

Error 5: Redundant Parameter Error  

Query: "a\_query\_that\_calls\_the\_tool (that only needs to fill in part of the parameters of the tool)"  

Function Calling Output:  
[
  \{\{
    "name": "name\_of\_the\_tool",
    "arguments": \{\{
      "parameter\_1": "parameter\_value",
      "parameter\_2": "parameter\_value"
    \}\}
  \}\}
]  

Error Message:  
\{\{
  "error": "RedundantParameter",
  "message": "The parameter 'parameter\_2' is not indicated by the query and should not be called."
\}\}

Thought of Error:  
Parameter 'parameter\_2' is unnecessary and was not specified in the query. Ensure only the required and specified parameters are included in the function call.

---

Error 6: Invalid Function Calling Output Format Error  

Query: "a\_query\_that\_calls\_the\_tool"  

Function Calling Output:  
\{\{
  "Name": "name\_of\_the\_tool",
  "Parameter": \{\{
    "parameter\_1": "parameter\_value",
    "parameter\_2": "parameter\_value"
  \}\}
\}\}

Error Message:  
\{\{
  "error": "InvalidFormat",
  "message": "The function calling output does not follow the required format and cannot be parsed."
\}\}

Thought of Error:  
The output format is incorrect due to improperly formatted keys and symbols. The correct function calling output should be:  
[
  \{\{
    "name": "func\_name1",
    "arguments": \{\{
      "parameter\_1": "value1",
      "parameter\_2": "value2"
    \}\}
  \}\}
]

---

Error 7: Redundant Information Error  

Query: "a\_query\_that\_calls\_the\_tool"  

Function Calling Output:  
"Based on the query, I will make a function call to the 'name\_of\_the\_tool' tool to get the query answered. Here is the output in the required JSON format: 
[
  \{\{
    'name': 'name\_of\_the\_tool',
    'arguments': \{\{
      'parameter\_1': 'parameter\_value',
      'parameter\_2': 'parameter\_value'
    \}\}
  \}\}
]"

Error Message:  
{{
  "error": "RedundantInformationError",
  "message": "The function calling output contains redundant text such as 'Based on the query, I will make a function call...' which is unnecessary."
}}

Thought of Error:  
No additional text should be included in the output. The correct function calling output should only contain:  
[
  \{\{
    "name": "func\_name1",
    "arguments": \{\{
      "parameter\_1": "value1",
      "parameter\_2": "value2"
    \}\}
  \}\}
]

------
% \end{tcolorbox}
% \begin{tcolorbox}[colback=bluee!10, colframe=bluee!80, title=Local Error Checklist Generation Prompt (cont'd)]

Instructions  

Now, generate an error checklist for the following tool:  

<tool\_info>

Note: You must strictly follow the format of the template.
\end{tcolorbox}

\begin{tcolorbox}
[colback=bluee!10, colframe=bluee!80, title=Negative Sample Generation Prompt]
\textbf{System Prompt:} You are provided with an error checklist, a tool calling query and its groundtruth answer. 

The error checklist of an example tool is as follows:

**Error Checklist Template**

Tool Information  

name: 'name\_of\_the\_tool'\ \\
(the same as the tool template in Local Error Checklist Generation Prompt)\\
---

Your task is to modify the groundtruth tool calling so that it fits one of the errors in the error checklist. For the Redundant Parameter Error, your generated redundant parameter should be one of the parameters in the tool information. If there is no extra parameter that can be chosen for Redundant Parameter Error, you can choose another errors. 

\#\#\#\#\# Note: DO NOT include not-exist parameters in your response, e.g., "extra\_param". \\
\#\#\#\#\# Note: You should return a modified response, for example: [\{"name": "getSocialEnterpriseInfo", "arguments": \{"enterprise\_name": "CommunityGrowth"\}\}]. \\
\#\#\#\#\# Note: Just provide the modified function calling output. DO NOT include other information". 

\textbf{User Prompt}: The user query is:\\
<user\_query>\\
The grountruth tool calling is:\\
<groundtruth>\\
Now please modify the groudtruth tool calling so that it meets one of the errors in the error checklist. Just return the modified tool calling. Do not explain your answer or include any other information. 
\end{tcolorbox}

\section{Implementation Details}
\label{app:imp_details}
The temperature for GPT-series base models is set as 0.2. The max\_new\_token is set as 512 when CoT is implemented, otherwise 256 for Llama-series base models. For other base models and parameters, we use the default setting as stated in the models' configuration files. The local error checklists and negative samples are generated using Qwen2.5-72B-Instruct. For the fine-tuning of Llama3-7B, Llama3.1-7B, Qwen2.5 series base models, we perform full-parameter tuning, while for Llama3-70B and Llama3.1-70B we perform LoRA \cite{hu2021lora}.

\section{Additional Experiment Results}
\label{app:exp}

\subsection{Additional Experiment Results on the Effectiveness of HiTEC-ICL}
Table \ref{tb:main_icl_full} presents the full results of experiments of tool calling performance with HiTEC-ICL.
\begin{table*}[]
\centering
\caption{Tool Calling Performance with HiTEC-ICL}
\fontsize{9}{10}\selectfont
\setlength{\tabcolsep}{2pt} % 调小列间距
\begin{tabular}{@{}cccccccccccc|cc@{}}
\toprule
\multicolumn{12}{c|}{Dataset (F1 Name $|$ F1 Name +  Parameter)}                                                                                                                                                                                                                                                                                                                                                                                                                                            & \multicolumn{2}{c}{F1 Average}                                                                  \\ \midrule
Model                                & Method                              & \multicolumn{2}{c}{\begin{tabular}[c]{@{}c@{}}API-Bank\\ L-1\end{tabular}}      & \multicolumn{2}{c}{\begin{tabular}[c]{@{}c@{}}API-Bank\\ L-2\end{tabular}}      & \multicolumn{2}{c}{Tool-Alpaca}                                                 & \multicolumn{2}{c}{\begin{tabular}[c]{@{}c@{}}Seal-Tools\\ (Single-Tool)\end{tabular}} & \multicolumn{2}{c|}{\begin{tabular}[c]{@{}c@{}}Nexus\\ Raven\end{tabular}}      & Name                                   & \begin{tabular}[c]{@{}c@{}}Name+\\ Param.\end{tabular} \\ \midrule
                                     & Vanilla                             & 85.26                                  & 77.06                                  & 78.87                                  & 61.02                                  & 86.23                                  & 56.47                                  & 94.97                                      & 87.43                                     & \textbf{91.42}                         & \textbf{81.27}                         & 87.35                                  & 72.65                                                  \\
                                     & + CoT                               & 65.28                                  & 59.13                                  & 59.2                                   & 55.53                                  & 75.46                                  & 58.43                                  & 89.91                                      & 80.71                                     & 83.92                                  & 75.69                                  & 74.75                                  & 65.90                                                  \\
                                     & + Function Calling                  & 85.57                                  & 82.47                                  & \textbf{88.72}                         & \textbf{65.87}                         & 80.00                                  & \textbf{72.73}                         & 96.10                                      & 88.50                                     & \textbackslash{}                       & \textbackslash{}                       & \textbackslash{}                       & \textbackslash{}                                       \\
\multirow{-4}{*}{GPT-3.5-Turbo-0125} & \cellcolor[HTML]{ECF4FF}+ HiTEC-ICL & \cellcolor[HTML]{ECF4FF}\textbf{92.38} & \cellcolor[HTML]{ECF4FF}\textbf{84.71} & \cellcolor[HTML]{ECF4FF}83.80          & \cellcolor[HTML]{ECF4FF}64.72          & \cellcolor[HTML]{ECF4FF}\textbf{87.27} & \cellcolor[HTML]{ECF4FF}58.79          & \cellcolor[HTML]{ECF4FF}\textbf{97.61}     & \cellcolor[HTML]{ECF4FF}\textbf{90.21}    & \cellcolor[HTML]{ECF4FF}91.22          & \cellcolor[HTML]{ECF4FF}81.08          & \cellcolor[HTML]{ECF4FF}\textbf{90.46} & \cellcolor[HTML]{ECF4FF}\textbf{75.90}                 \\ \midrule
                                     & Vanilla                             & 94.77                                  & 86.66                                  & 71.09                                  & 70.43                                  & 87.12                                  & 60.75                                  & 94.86                                      & 88.55                                     & \textbf{94.13}                         & 82.92                                  & 88.39                                  & 77.86                                                  \\
                                     & + CoT                               & 91.70                                  & 84.84                                  & 67.86                                  & 65.74                                  & 80.95                                  & 56.46                                  & 91.50                                      & 86.07                                     & 92.18                                  & \textbf{85.00}                         & 84.84                                  & 75.62                                                  \\
                                     & + Function Calling                  & 84.03                                  & 81.40                                  & \textbf{78.49}                         & 62.32                                  & 87.11                                  & \textbf{62.16}                         & 84.94                                      & 84.88                                     & \textbackslash{}                       & \textbackslash{}                       & \textbackslash{}                       & \textbackslash{}                                       \\
\multirow{-4}{*}{GPT-4-Turbo}        & \cellcolor[HTML]{ECF4FF}+ HiTEC-ICL & \cellcolor[HTML]{ECF4FF}\textbf{94.96} & \cellcolor[HTML]{ECF4FF}\textbf{88.44} & \cellcolor[HTML]{ECF4FF}71.21          & \cellcolor[HTML]{ECF4FF}\textbf{72.68} & \cellcolor[HTML]{ECF4FF}\textbf{87.35} & \cellcolor[HTML]{ECF4FF}61.95          & \cellcolor[HTML]{ECF4FF}\textbf{96.00}     & \cellcolor[HTML]{ECF4FF}\textbf{89.37}    & \cellcolor[HTML]{ECF4FF}92.81          & \cellcolor[HTML]{ECF4FF}83.15          & \cellcolor[HTML]{ECF4FF}\textbf{88.47} & \cellcolor[HTML]{ECF4FF}\textbf{79.12}                 \\ \midrule
                                     & Vanilla                             & 69.19                                  & 67.73                                  & 75.54                                  & 48.91                                  & 40.46                                  & 24.86                                  & 86.54                                      & 82.51                                     & 14.83                                  & 17.44                                  & 57.31                                  & 48.29                                                  \\
                                     & + CoT                               & 43.21                                  & 45.05                                  & 52.31                                  & 45.65                                  & 60.58                                  & 42.07                                  & 76.89                                      & 75.16                                     & 64.44                                  & 52.22                                  & 59.49                                  & 52.03                                                  \\
\multirow{-3}{*}{Llama3-8B}          & \cellcolor[HTML]{ECF4FF}+ HiTEC-ICL & \cellcolor[HTML]{ECF4FF}\textbf{75.93} & \cellcolor[HTML]{ECF4FF}\textbf{70.55} & \cellcolor[HTML]{ECF4FF}\textbf{77.30} & \cellcolor[HTML]{ECF4FF}\textbf{56.01} & \cellcolor[HTML]{ECF4FF}\textbf{85.92} & \cellcolor[HTML]{ECF4FF}\textbf{55.49} & \cellcolor[HTML]{ECF4FF}\textbf{94.40}     & \cellcolor[HTML]{ECF4FF}\textbf{87.27}    & \cellcolor[HTML]{ECF4FF}\textbf{82.07} & \cellcolor[HTML]{ECF4FF}\textbf{58.30} & \cellcolor[HTML]{ECF4FF}\textbf{83.12} & \cellcolor[HTML]{ECF4FF}\textbf{65.52}                 \\ \midrule
                                     & Vanilla                             & \textbf{86.47}                         & 78.87                                  & 66.67                                  & 63.65                                  & 49.64                                  & 47.73                                  & 88.00                                      & \textbf{86.81}                            & 75.43                                  & 73.25                                  & 73.24                                  & 70.06                                                  \\
                                     & + CoT                               & 78.41                                  & 74.52                                  & 63.04                                  & 52.53                                  & 80.75                                  & 52.61                                  & 87.03                                      & 81.25                                     & 78.43                                  & 72.24                                  & 77.53                                  & 66.63                                                  \\
\multirow{-3}{*}{Llama3-70B}         & \cellcolor[HTML]{ECF4FF}+ HiTEC-ICL & \cellcolor[HTML]{ECF4FF}84.85          & \cellcolor[HTML]{ECF4FF}\textbf{78.97} & \cellcolor[HTML]{ECF4FF}\textbf{79.05} & \cellcolor[HTML]{ECF4FF}\textbf{63.88} & \cellcolor[HTML]{ECF4FF}\textbf{86.76} & \cellcolor[HTML]{ECF4FF}\textbf{58.62} & \cellcolor[HTML]{ECF4FF}\textbf{95.55}     & \cellcolor[HTML]{ECF4FF}86.73             & \cellcolor[HTML]{ECF4FF}\textbf{85.40} & \cellcolor[HTML]{ECF4FF}\textbf{74.95} & \cellcolor[HTML]{ECF4FF}\textbf{86.32} & \cellcolor[HTML]{ECF4FF}\textbf{72.63}                 \\ \midrule
                                     & Vanilla                             & 73.06                                  & 63.62                                  & 78.23                                  & 50.25                                  & 84.06                                  & 50.79                                  & 94.97                                      & 86.53                                     & \textbf{87.93}                         & 72.17                                  & 83.65                                  & 64.67                                                  \\
                                     & + CoT                               & 65.22                                  & 61.61                                  & 75.54                                  & 48.52                                  & 63.31                                  & 42.57                                  & 88.55                                      & 80.36                                     & 76.71                                  & 67.32                                  & 73.87                                  & 60.08                                                  \\
\multirow{-3}{*}{Llama3.1-8B}        & \cellcolor[HTML]{ECF4FF}+ HiTEC-ICL & \cellcolor[HTML]{ECF4FF}\textbf{75.07} & \cellcolor[HTML]{ECF4FF}\textbf{64.61} & \cellcolor[HTML]{ECF4FF}\textbf{82.23} & \cellcolor[HTML]{ECF4FF}\textbf{51.29} & \cellcolor[HTML]{ECF4FF}\textbf{85.92} & \cellcolor[HTML]{ECF4FF}\textbf{55.94} & \cellcolor[HTML]{ECF4FF}\textbf{96.95}     & \cellcolor[HTML]{ECF4FF}\textbf{87.94}    & \cellcolor[HTML]{ECF4FF}87.76          & \cellcolor[HTML]{ECF4FF}\textbf{72.35} & \cellcolor[HTML]{ECF4FF}\textbf{85.59} & \cellcolor[HTML]{ECF4FF}\textbf{66.42}                 \\ \midrule
                                     & Vanilla                             & 90.15                                  & 76.42                                  & \textbf{80.34}                         & 62.20                                  & \textbf{86.03}                         & 55.83                                  & \textbf{97.42}                             & 86.19                                     & 93.75                                  & 82.72                                  & 89.54                                  & 72.67                                                  \\
                                     & + CoT                               & 87.05                                  & 73.64                                  & 80.31                                  & 60.15                                  & 83.90                                  & 55.38                                  & 95.16                                      & 85.49                                     & 91.64                                  & 78.73                                  & 87.61                                  & 70.68                                                  \\
\multirow{-3}{*}{Llama3.1-70B}       & \cellcolor[HTML]{ECF4FF}+ HiTEC-ICL & \cellcolor[HTML]{ECF4FF}\textbf{93.21} & \cellcolor[HTML]{ECF4FF}\textbf{79.05} & \cellcolor[HTML]{ECF4FF}80.18          & \cellcolor[HTML]{ECF4FF}\textbf{64.50} & \cellcolor[HTML]{ECF4FF}84.96          & \cellcolor[HTML]{ECF4FF}\textbf{56.83} & \cellcolor[HTML]{ECF4FF}95.86              & \cellcolor[HTML]{ECF4FF}\textbf{87.19}    & \cellcolor[HTML]{ECF4FF}\textbf{94.54} & \cellcolor[HTML]{ECF4FF}\textbf{83.44} & \cellcolor[HTML]{ECF4FF}\textbf{89.75} & \cellcolor[HTML]{ECF4FF}\textbf{74.20}                 \\ \bottomrule
\end{tabular}
\label{tb:main_icl_full}
\end{table*}

\subsection{Additional Experiment Results on the Effectiveness of HiTEC-KTO}
Table \ref{tb:main_kto_full} presents the full results of experiments of tool calling performance with HiTEC-KTO.
\begin{table*}[]
\caption{Tool Calling Performance with HiTEC-KTO}
\fontsize{9}{10}\selectfont
\setlength{\tabcolsep}{3pt} % 调小列间距
\begin{tabular}{@{}ccccccccccccc|cc@{}}
\toprule
\multicolumn{13}{c|}{Dataset (F1 Name $|$ F1 Name + Parameter)}                                                                                                                                                                                                                                                                                                                                                                                                                                                                                                                                                                                                           & \multicolumn{2}{c}{F1 Average}                                                                   \\ \midrule
\multicolumn{1}{c|}{\begin{tabular}[c]{@{}c@{}}Model\\ Series\end{tabular}}                      & Method                                                                                         & Model                                & \multicolumn{2}{c}{\begin{tabular}[c]{@{}c@{}}API-Bank\\ L-1\end{tabular}}      & \multicolumn{2}{c}{\begin{tabular}[c]{@{}c@{}}API-Bank\\ L-2\end{tabular}}      & \multicolumn{2}{c}{Tool-Alpaca}                                                 & \multicolumn{2}{c}{\begin{tabular}[c]{@{}c@{}}Seal-Tools\\ (Single-Tool)\end{tabular}} & \multicolumn{2}{c|}{\begin{tabular}[c]{@{}c@{}}Nexus\\ Raven\end{tabular}}      & Name                                   & \begin{tabular}[c]{@{}c@{}}Name +\\ Param.\end{tabular} \\ \midrule
\multicolumn{1}{c|}{}                                                                            &                                                                                                & Llama3-8B                            & 69.19                                  & 67.73                                  & 75.54                                  & 48.91                                  & 40.46                                  & 24.86                                  & 86.54                                      & \textbf{82.51}                            & 14.83                                  & 17.44                                  & 57.31                                  & 48.29                                                   \\
\multicolumn{1}{c|}{}                                                                            & \multirow{-2}{*}{Baseline}                                                                     & Llama3-70B                           & 86.47                                  & 78.87                                  & 66.67                                  & \textbf{63.65}                         & 49.64                                  & 47.73                                  & 88.00                                      & 86.81                                     & 75.43                                  & 73.25                                  & 73.24                                  & 70.06                                                   \\ \cmidrule(l){2-15} 
\multicolumn{1}{c|}{}                                                                            & \cellcolor[HTML]{ECF4FF}                                                                       & \cellcolor[HTML]{ECF4FF}Llama3-8B    & \cellcolor[HTML]{ECF4FF}\textbf{94.99} & \cellcolor[HTML]{ECF4FF}\textbf{84.99} & \cellcolor[HTML]{ECF4FF}\textbf{90.42} & \cellcolor[HTML]{ECF4FF}\textbf{64.20} & \cellcolor[HTML]{ECF4FF}\textbf{87.02} & \cellcolor[HTML]{ECF4FF}\textbf{52.40} & \cellcolor[HTML]{ECF4FF}\textbf{96.43}     & \cellcolor[HTML]{ECF4FF}81.93             & \cellcolor[HTML]{ECF4FF}\textbf{78.17} & \cellcolor[HTML]{ECF4FF}\textbf{64.60} & \cellcolor[HTML]{ECF4FF}\textbf{89.41} & \cellcolor[HTML]{ECF4FF}\textbf{69.62}                  \\
\multicolumn{1}{c|}{\multirow{-4}{*}{\begin{tabular}[c]{@{}c@{}}Llama3\\ Series\end{tabular}}}   & \multirow{-2}{*}{\cellcolor[HTML]{ECF4FF}\begin{tabular}[c]{@{}c@{}}HiTEC-\\ KTO\end{tabular}} & \cellcolor[HTML]{ECF4FF}Llama3-70B   & \cellcolor[HTML]{ECF4FF}\textbf{86.88} & \cellcolor[HTML]{ECF4FF}\textbf{79.71} & \cellcolor[HTML]{ECF4FF}\textbf{75.61} & \cellcolor[HTML]{ECF4FF}63.08          & \cellcolor[HTML]{ECF4FF}\textbf{91.17} & \cellcolor[HTML]{ECF4FF}\textbf{60.83} & \cellcolor[HTML]{ECF4FF}\textbf{96.47}     & \cellcolor[HTML]{ECF4FF}\textbf{88.37}    & \cellcolor[HTML]{ECF4FF}\textbf{91.85} & \cellcolor[HTML]{ECF4FF}\textbf{81.92} & \cellcolor[HTML]{ECF4FF}\textbf{88.40} & \cellcolor[HTML]{ECF4FF}\textbf{74.78}                  \\ \midrule
\multicolumn{1}{c|}{}                                                                            &                                                                                                & Llama3.1-8B                          & 73.06                                  & 63.62                                  & 78.23                                  & 50.25                                  & 84.06                                  & 50.79                                  & 94.97                                      & \textbf{86.53}                            & 87.93                                  & 72.17                                  & 83.65                                  & 64.67                                                   \\
\multicolumn{1}{c|}{}                                                                            & \multirow{-2}{*}{Baseline}                                                                     & Llama3.1-70B                         & 90.15                                  & 76.42                                  & 80.34                                  & 62.20                                  & 86.03                                  & 55.83                                  & \textbf{97.42}                             & 86.19                                     & 93.75                                  & 82.72                                  & 89.54                                  & 72.67                                                   \\ \cmidrule(l){2-15} 
\multicolumn{1}{c|}{}                                                                            & \cellcolor[HTML]{ECF4FF}                                                                       & \cellcolor[HTML]{ECF4FF}Llama3.1-8B  & \cellcolor[HTML]{ECF4FF}\textbf{87.47} & \cellcolor[HTML]{ECF4FF}\textbf{80.99} & \cellcolor[HTML]{ECF4FF}\textbf{85.61} & \cellcolor[HTML]{ECF4FF}\textbf{61.02} & \cellcolor[HTML]{ECF4FF}\textbf{84.18} & \cellcolor[HTML]{ECF4FF}\textbf{56.45} & \cellcolor[HTML]{ECF4FF}\textbf{94.44}     & \cellcolor[HTML]{ECF4FF}86.18             & \cellcolor[HTML]{ECF4FF}\textbf{89.98} & \cellcolor[HTML]{ECF4FF}\textbf{82.28} & \cellcolor[HTML]{ECF4FF}\textbf{88.34} & \cellcolor[HTML]{ECF4FF}\textbf{73.38}                  \\
\multicolumn{1}{c|}{\multirow{-4}{*}{\begin{tabular}[c]{@{}c@{}}Llama3.1\\ Series\end{tabular}}} & \multirow{-2}{*}{\cellcolor[HTML]{ECF4FF}\begin{tabular}[c]{@{}c@{}}HiTEC-\\ KTO\end{tabular}} & \cellcolor[HTML]{ECF4FF}Llama3-70B   & \cellcolor[HTML]{ECF4FF}\textbf{90.78} & \cellcolor[HTML]{ECF4FF}\textbf{77.07} & \cellcolor[HTML]{ECF4FF}\textbf{86.44} & \cellcolor[HTML]{ECF4FF}\textbf{65.14} & \cellcolor[HTML]{ECF4FF}\textbf{86.67} & \cellcolor[HTML]{ECF4FF}\textbf{57.32} & \cellcolor[HTML]{ECF4FF}\textbf{98.14}     & \cellcolor[HTML]{ECF4FF}\textbf{90.01}    & \cellcolor[HTML]{ECF4FF}\textbf{94.84} & \cellcolor[HTML]{ECF4FF}\textbf{82.87} & \cellcolor[HTML]{ECF4FF}\textbf{91.37} & \cellcolor[HTML]{ECF4FF}\textbf{74.48}                  \\ \midrule
\multicolumn{1}{c|}{}                                                                            &                                                                                                & Hammer2-0.5B                         & 71.20                                  & 59.03                                  & 43.32                                  & 38.22                                  & 64.46                                  & \textbf{41.83}                         & 93.86                                      & 83.03                                     & 64.72                                  & 45.52                                  & 67.51                                  & 53.53                                                   \\
\multicolumn{1}{c|}{}                                                                            &                                                                                                & Hammer2-1.5B                         & 88.63                                  & 79.26                                  & 80.51                                  & \textbf{62.82}                         & 80.74                                  & 51.88                                  & 96.10                                      & 87.16                                     & 85.85                                  & 63.76                                  & 86.37                                  & 68.98                                                   \\
\multicolumn{1}{c|}{}                                                                            &                                                                                                & Hammer2-3B                           & 88.63                                  & 79.04                                  & 77.11                                  & \textbf{57.58}                         & 78.23                                  & 53.28                                  & 93.08                                      & 85.60                                     & \textbf{89.14}                         & \textbf{66.71}                         & 85.24                                  & 68.44                                                   \\
\multicolumn{1}{c|}{}                                                                            & \multirow{-4}{*}{Baseline}                                                                     & Hammer2-7B                           & 88.91                                  & 81.28                                  & 75.96                                  & 58.36                                  & 81.74                                  & 57.07                                  & 94.62                                      & 87.84                                     & 90.76                                  & \textbf{80.96}                         & 86.40                                  & 73.10                                                   \\ \cmidrule(l){2-15} 
\multicolumn{1}{c|}{}                                                                            & \cellcolor[HTML]{ECF4FF}                                                                       & \cellcolor[HTML]{ECF4FF}Qwen2.5-0.5B & \cellcolor[HTML]{ECF4FF}\textbf{88.29} & \cellcolor[HTML]{ECF4FF}\textbf{78.34} & \cellcolor[HTML]{ECF4FF}\textbf{81.76} & \cellcolor[HTML]{ECF4FF}\textbf{54.52} & \cellcolor[HTML]{ECF4FF}\textbf{73.54} & \cellcolor[HTML]{ECF4FF}38.76          & \cellcolor[HTML]{ECF4FF}\textbf{96.80}     & \cellcolor[HTML]{ECF4FF}\textbf{88.18}    & \cellcolor[HTML]{ECF4FF}\textbf{82.83} & \cellcolor[HTML]{ECF4FF}\textbf{63.89} & \cellcolor[HTML]{ECF4FF}\textbf{84.64} & \cellcolor[HTML]{ECF4FF}\textbf{64.74}                  \\
\multicolumn{1}{c|}{}                                                                            & \cellcolor[HTML]{ECF4FF}                                                                       & \cellcolor[HTML]{ECF4FF}Qwen2.5-1.5B & \cellcolor[HTML]{ECF4FF}\textbf{88.92} & \cellcolor[HTML]{ECF4FF}\textbf{79.46} & \cellcolor[HTML]{ECF4FF}\textbf{81.36} & \cellcolor[HTML]{ECF4FF}60.66          & \cellcolor[HTML]{ECF4FF}\textbf{82.27} & \cellcolor[HTML]{ECF4FF}\textbf{52.26} & \cellcolor[HTML]{ECF4FF}\textbf{96.99}     & \cellcolor[HTML]{ECF4FF}\textbf{89.75}    & \cellcolor[HTML]{ECF4FF}\textbf{86.37} & \cellcolor[HTML]{ECF4FF}\textbf{64.40} & \cellcolor[HTML]{ECF4FF}\textbf{87.18} & \cellcolor[HTML]{ECF4FF}\textbf{69.31}                  \\
\multicolumn{1}{c|}{}                                                                            & \cellcolor[HTML]{ECF4FF}                                                                       & \cellcolor[HTML]{ECF4FF}Qwen2.5-3B   & \cellcolor[HTML]{ECF4FF}\textbf{89.07} & \cellcolor[HTML]{ECF4FF}\textbf{80.16} & \cellcolor[HTML]{ECF4FF}\textbf{85.51} & \cellcolor[HTML]{ECF4FF}54.61          & \cellcolor[HTML]{ECF4FF}\textbf{85.42} & \cellcolor[HTML]{ECF4FF}\textbf{56.08} & \cellcolor[HTML]{ECF4FF}\textbf{96.28}     & \cellcolor[HTML]{ECF4FF}\textbf{89.51}    & \cellcolor[HTML]{ECF4FF}83.05          & \cellcolor[HTML]{ECF4FF}62.35          & \cellcolor[HTML]{ECF4FF}\textbf{87.87} & \cellcolor[HTML]{ECF4FF}\textbf{68.54}                  \\
\multicolumn{1}{c|}{\multirow{-8}{*}{\begin{tabular}[c]{@{}c@{}}Qwen2.5\\ Series\end{tabular}}}  & \multirow{-4}{*}{\cellcolor[HTML]{ECF4FF}\begin{tabular}[c]{@{}c@{}}HiTEC-\\ KTO\end{tabular}} & \cellcolor[HTML]{ECF4FF}Qwen2.5-7B   & \cellcolor[HTML]{ECF4FF}\textbf{89.38} & \cellcolor[HTML]{ECF4FF}\textbf{81.67} & \cellcolor[HTML]{ECF4FF}\textbf{88.00} & \cellcolor[HTML]{ECF4FF}\textbf{59.93} & \cellcolor[HTML]{ECF4FF}\textbf{87.63} & \cellcolor[HTML]{ECF4FF}\textbf{59.19} & \cellcolor[HTML]{ECF4FF}\textbf{96.27}     & \cellcolor[HTML]{ECF4FF}\textbf{89.29}    & \cellcolor[HTML]{ECF4FF}\textbf{91.70} & \cellcolor[HTML]{ECF4FF}78.17          & \cellcolor[HTML]{ECF4FF}\textbf{90.60} & \cellcolor[HTML]{ECF4FF}\textbf{73.65}                  \\ \bottomrule
\end{tabular}
\label{tb:main_kto_full}
\end{table*}

\subsection{Additional Experiment Results on Ablation Study on The Hierarchical Error Checklist}
Table \ref{tb:ablation} and \ref{tb:ablation_kto_full} present the full results of ablation study on the hierarchical error checklist.

\begin{table*}[]
\centering
\caption{Ablation Study on The Hierarchical Error Checklist of HiTEC-ICL}
\fontsize{9}{10}\selectfont
\setlength{\tabcolsep}{3pt} % 调小列间距
\begin{tabular}{@{}cccccccccc|cc@{}}
\toprule
\multicolumn{10}{c|}{Base Model (F1 Name | F1 Name + Parameter)}                                                                                                                                                                         & \multicolumn{2}{c}{F1 Average}                                          \\ \midrule
Dataset                      & Method         & \multicolumn{2}{c}{\begin{tabular}[c]{@{}c@{}}GPT-3.5-\\ Turbo-0125\end{tabular}} & \multicolumn{2}{c}{GPT-4-Turbo} & \multicolumn{2}{c}{Llama3-8B}   & \multicolumn{2}{c|}{Llama3.1-8B} & Name           & \begin{tabular}[c]{@{}c@{}}Name+\\ Param.\end{tabular} \\ \midrule
\multirow{3}{*}{Tool Alpaca} & HiTEC-ICL      & \textbf{88.09}                          & \textbf{58.11}                          & \textbf{87.35} & \textbf{61.95} & \textbf{85.92} & \textbf{55.49} & 85.92           & \textbf{55.94} & \textbf{86.82} & \textbf{57.87}                                         \\
                             & w/o Local EC   & 87.05                                   & 57.98                                   & 86.36          & 60.67          & 82.71          & 52.39          & \textbf{86.43}  & 52.33          & 85.64          & 55.84                                                  \\
                             & w/o Glb-loc EC & 86.23                                   & 56.47                                   & 87.12          & 60.75          & 40.46          & 24.86          & 84.06           & 50.79          & 74.47          & 48.22                                                  \\ \midrule
\multirow{3}{*}{Seal-Tools}  & HiTEC-ICL      & 97.61                                   & \textbf{90.21}                          & \textbf{96.00} & \textbf{89.37} & \textbf{94.40} & \textbf{87.27} & \textbf{96.95}  & \textbf{87.94} & \textbf{96.24} & \textbf{88.70}                                         \\
                             & w/o Local EC   & \textbf{97.62}                          & 90.11                                   & 95.99          & 88.82          & 94.22          & 83.06          & 95.43           & 85.66          & 95.81          & 86.91                                                  \\
                             & w/o Glb-loc EC & 94.97                                   & 87.43                                   & 94.86          & 88.55          & 86.54          & 82.51          & 94.97           & 86.53          & 92.84          & 86.26                                                  \\ \bottomrule
\end{tabular}
\label{tb:ablation}
\end{table*}

\begin{table*}[]
\centering
\caption{Ablation Study on The Hierarchical Error Checklist of HiTEC-KTO}
\vspace{-0.3cm}
\fontsize{9}{10}\selectfont
\setlength{\tabcolsep}{2.5pt} % 调小列间距
\begin{tabular}{@{}cccccccccc|cc@{}}
\toprule
\multicolumn{10}{c|}{Base Model (F1 Name | F1 Name + Parameter)}                                                                                                                        & \multicolumn{2}{c}{F1 Average}                                          \\ \midrule
Dataset                      & Method         & \multicolumn{2}{c}{Qwen2.5-0.5B} & \multicolumn{2}{c}{Qwen2.5-1.5B} & \multicolumn{2}{c}{Qwen2.5-3B}  & \multicolumn{2}{c|}{Qwen2.5-7B} & Name           & \begin{tabular}[c]{@{}c@{}}Name+\\ Param.\end{tabular} \\ \midrule
\multirow{3}{*}{Tool Alpaca} & HiTEC-KTO      & \textbf{73.54}  & \textbf{38.76} & \textbf{82.27}  & \textbf{52.26} & \textbf{85.42} & \textbf{56.08} & 87.63 & \textbf{59.19} & \textbf{82.22} & \textbf{51.57}                                         \\
                             & w/o Local EC   & 57.51           & 31.71          & 60.61           & 28.97          & 77.62          & 51.61          & \textbf{87.86} & 58.40          & 70.90          & 42.67                                                  \\
                             & w/o Glb-loc EC & 0.79            & 0.83           & 22.83           & 16.03          & 73.45          & 47.75          & 74.35          & 57.12          & 42.86          & 30.43                                                  \\ \midrule
\multirow{3}{*}{Seal-Tools}  & HiTEC-KTO      & \textbf{96.80}  & \textbf{88.18} & \textbf{96.99}  & \textbf{89.75} & \textbf{96.28} & \textbf{89.51} & \textbf{96.27} & \textbf{89.29} & \textbf{96.59} & \textbf{89.18}                                         \\
                             & w/o Local EC   & 86.47           & 77.57          & 87.75           & 68.99          & 92.75          & 82.60          & 94.17          & 86.66          & 90.44          & 78.96                                                  \\
                             & w/o Glb-loc EC & 1.06            & 1.90           & 41.86           & 37.34          & 83.98          & 77.99          & 89.80          & 84.87          & 54.18          & 50.53                                                  \\ \bottomrule
\end{tabular}
\label{tb:ablation_kto_full}
\vspace{-0.3cm}
\end{table*}

\subsection{Additional Experiment Results on Ablation Study on Error Types}
\begin{figure*}[htbp]
    \centering
    \subfigure[Tool Alpaca - F1 Name]{
        \centering
        \includegraphics[width=0.25\linewidth]{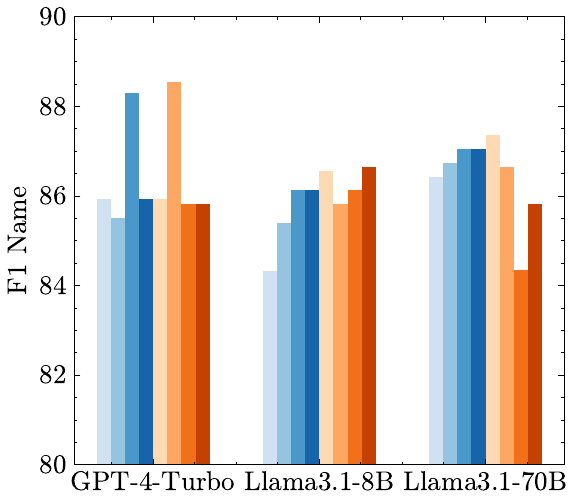}
    }%
    \subfigure[Tool Alpaca - F1 Name + Parameter]{
        \centering
        \includegraphics[width=0.25\linewidth]{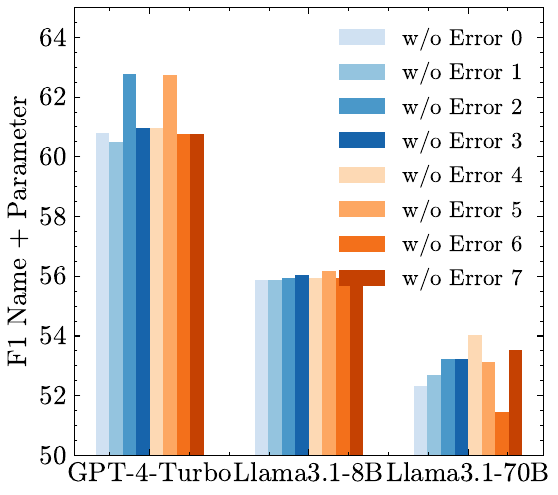}
   }%
    \subfigure[Seal-Tools - F1 Name]{
        \centering
        \includegraphics[width=0.25\linewidth]{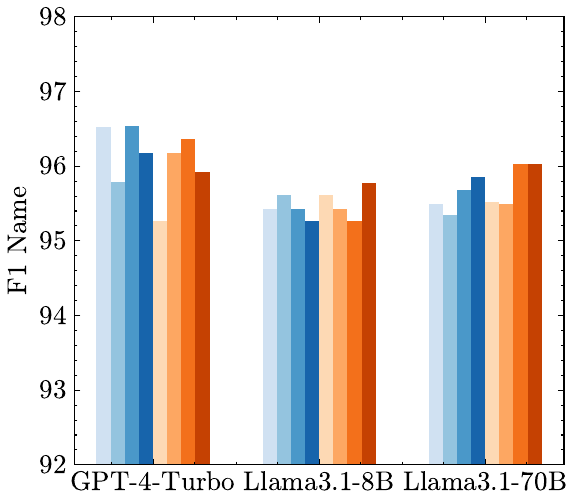}
    }%
    \subfigure[Seal-Tools - F1 Name + Parameter]{
        \centering
        \includegraphics[width=0.25\linewidth]{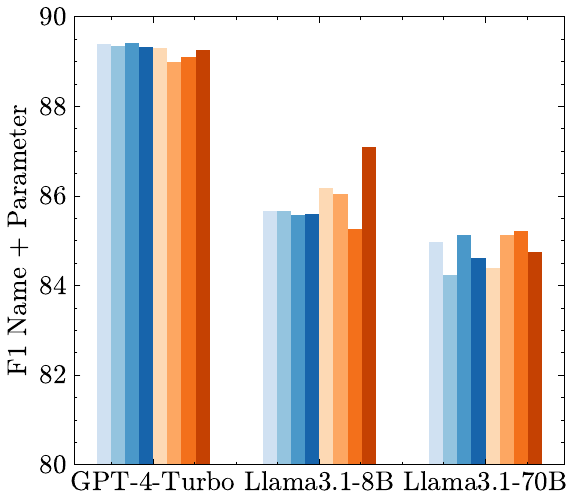}
        }
    \caption{Ablation Study on Error Types}
    \label{fig:error_type}
\end{figure*}

Figure \ref{fig:error_type} illustrates the full results on ablation study on error types.

\begin{tcolorbox}[colback=bluee!10, colframe=bluee!80, title=Example: Small Difference between Positive and Negative Answer]
\label{example:small_diff}
        \textbf{"user"}: "You are a tool calling assistant. In order to complete the user"s request, you need to select one or more appropriate tools from the following tools and fill in the correct values for the tool parameters. Your specific tasks are:\\
        1. Make one or more function/tool calls to meet the request based on the question.\\
        2. If none of the function can be used, point it out and refuse to answer.
        3. If the given question lacks the parameters required by the function, also point it out.\\

        [\{"name": "polygon\_area\_shoelace", "description": "Calculates the area of a polygon using the shoelace formula.", "parameters": \{"vertices": \{"description": "A list of polygon vertices represented as tuples (x, y).", "type": "List[Tuple[float, float]]"\}\}\}, \{"name": "find\_n\_largest\_numbers", "description": "Finds the n largest numbers in a list.", "parameters": \{"nums": \{"description": "The list of numbers.", "type": "List[int]"\}, "n": \{"description": "The number of largest numbers to find.", "type": "int"\}\}\}]

        The output MUST strictly adhere to the following JSON format, and NO other text MUST be included.
        The example format is as follows. Please make sure the parameter type is correct. If no function call is needed, please directly output an empty list "[]"

        [
            \{"name": "func\_name1", "arguments": \{"argument1": "value1", "argument2": "value2"\}\},
            ... (more tool calls as required)
            ]

        What is the area of a hexagon with vertices at (1, 1), (5, 1), (7, 5), (5, 9), (1, 9), and (0, 5)? Also, extract the 4 largest numbers in the list [120, 130, 140, 150, 160]"\\

        \textbf{"chosen"}: 
        [\{"name": "polygon\_area\_shoelace", "arguments": \{"vertices": [[1, 1], [5, 1], [7, 5], [5, 9], [1, 9], [0, 5]]\}\}, \{"name": "find\_n\_largest\_numbers", "arguments": \{"nums": [120, 130, 140, 150, 160], "n": 4\}\}],\\

        \textbf{"rejected"}: [\{"name": "polygon\_area\_shoelace", "arguments": \{"vertices": [[1, 1], [5, 1], [7, 5], [5, 9], [1, 9], [0, 5]], "n": 4\}\},
        \{"name": "find\_n\_largest\_numbers", "arguments": \{"nums": [120, 130, 140, 150, 160], "n": 4\}\}
        ]
\end{tcolorbox}

\section{The Comparison Between DPO and KTO}
\label{appendix:dpo_kto}

In our PTC dataset, the positive and negative responses are often similar. This similarity arises because errors in negative responses typically occur when assigning specific values to parameters. For instance, as demonstrated in Example~\ref{example:small_diff}, the only difference between positive and negative responses is the parameter $n: 4$. In the following analysis, we provide a detailed explanation of why this difference leads to the DPO's failure mode.

To provide a brief analysis of why DPO fails, consider the DPO objective:

\begin{equation}\mathcal{L}_{DPO}(x, y_w, y_l; \theta)
    = -\log \sigma\left(\beta r_{\theta}(x, y_w)-\beta  r_{\theta}(x, y_l)\right),
\end{equation}
where $(x, y_w, y_l)$ are paired data with $x$ as the prompt, $y_w$ as the positive answer and $y_l$ as the negative answer. 
$r_{\theta}(x, y)$ is the log-ratio of the likelihoods of answer $y$ between the training model (i.e., $\pi_{\theta}(y|x)$) and the reference model (i.e., $\pi_{\text{ref}}(y|x)$), where $\theta$ is the model parameter. $r_{\theta}(x, y)$ is denoted as $r_{\theta}(x, y) = \frac{\pi_{\theta}(y|x)}{\pi_{\text{ref}}(y|x)}$.

\begin{figure}
    \centering
    \subfigure[Training Loss]{
    \centering
        \includegraphics[width=0.225\linewidth]{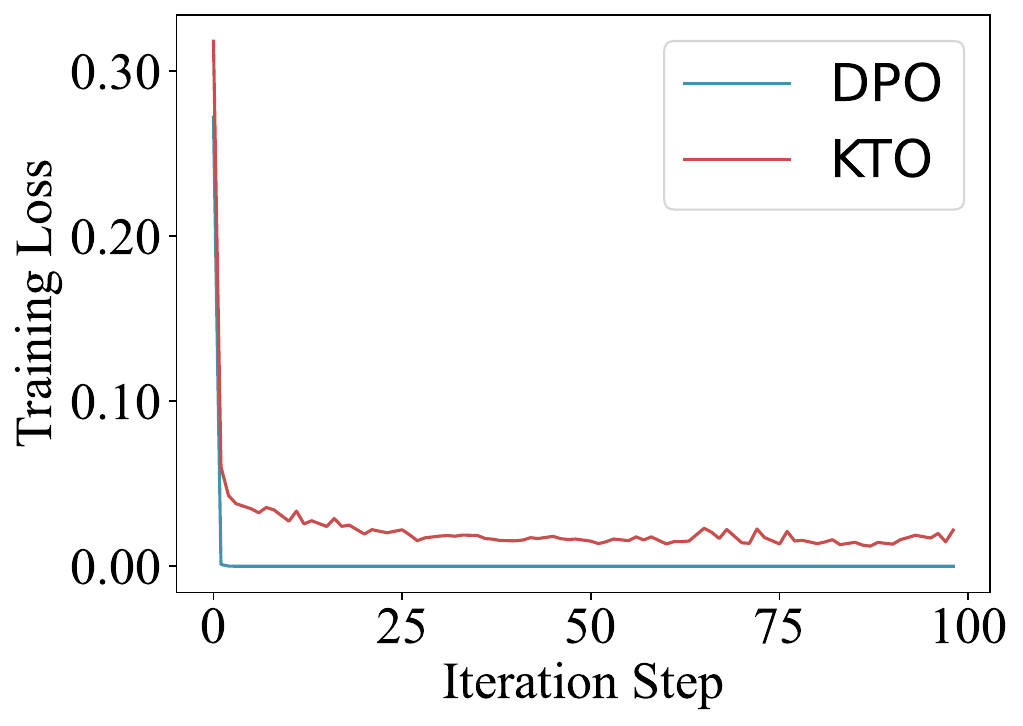}
        \label{subfig:training_loss}
    }
    \subfigure[Gradient Norm]{
    \centering
        \includegraphics[width=0.225\linewidth]{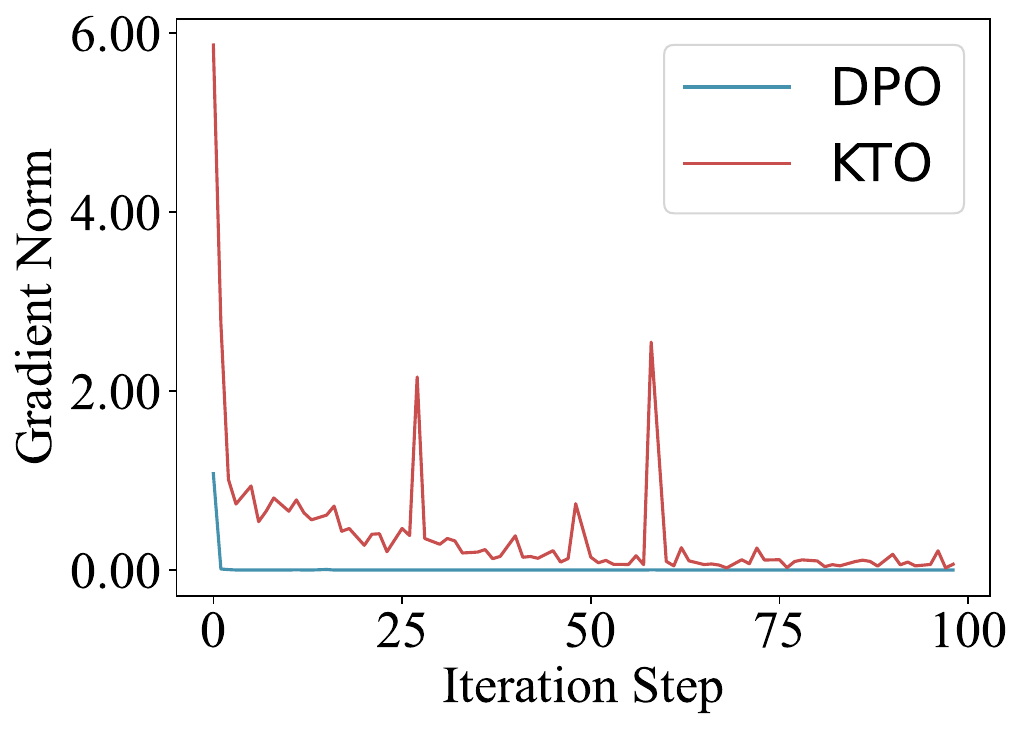}
         \label{subfig:gradient_norm}
    }
    \subfigure[Logps of Chosen]{
    \centering
        \includegraphics[width=0.225\linewidth]{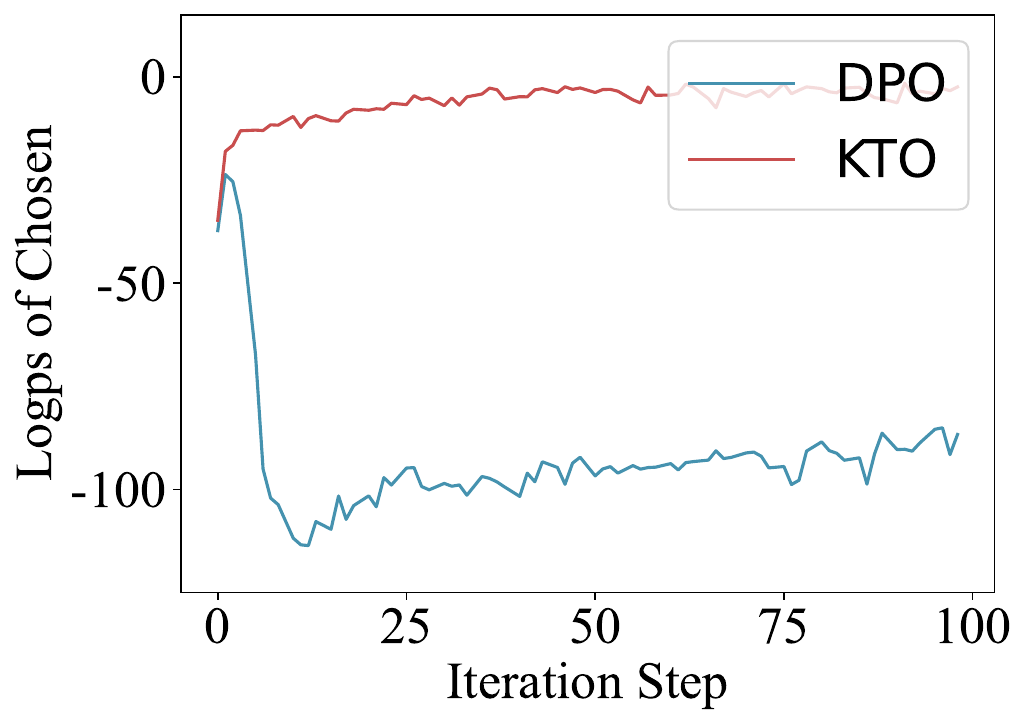}
         \label{subfig:logps_chosen}
    }
    \subfigure[Logps of Rejected]{
    \centering
        \includegraphics[width=0.225\linewidth]{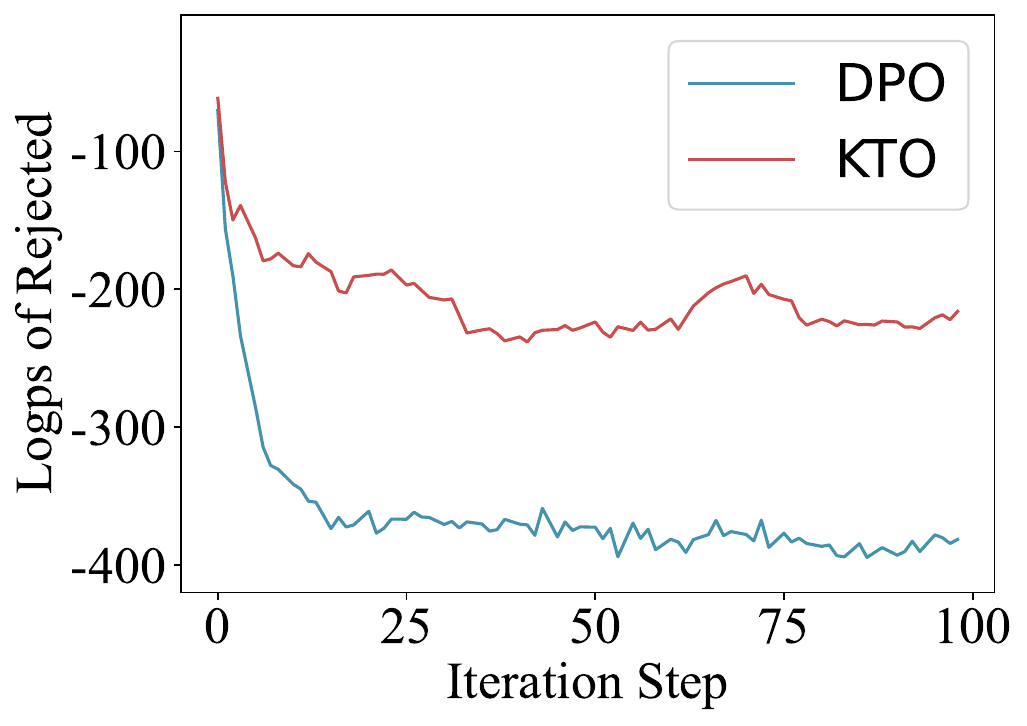}
        \label{subfig:logps_rejected}
    }
\end{figure}

Suppose $y_w$ and $y_l$ are length $K$ sequences, and they only differ at i-th token, i.e., $y_w= [t_1, \cdots, t_{i-1}, t_{i}^w, t_{i+1}, \cdots, t_K]$, and $y_l= [t_1, \cdots, t_{i-1}, t_{i}^l, t_{i+1}, \cdots, t_K]$. Then
the gradient of $\mathcal{L}_{DPO}$ is calculated as:

\begin{equation}
\begin{split}
    \nabla_{\theta}\mathcal{L}_{DPO} 
 =& -\beta \cdot \sigma\left(-c\right)\cdot \nabla_{\theta}\left[\log\pi_{\theta}(y_w|x) -\log\pi_{\theta}(y_w|x)\right]\\
 \approx&-\beta \cdot \sigma\left(-c\right)\cdot \nabla_{\theta}\left[\log\pi_{\theta}(t_{i}^w, y^{> i}|x, y^{<i}) -\log\pi_{\theta}( t_{i}^l, y^{> i}|x, y^{<i})\right],
\end{split}
\end{equation}
where $c=\beta r_{\theta}(x, y_w)-\beta  r_{\theta}(x, y_l)$, $y^{<i}= [t_1, \cdots, t_{i-1}]$ is the sequence before i-th token, and similarly, $y^{\geq i}$ is the sequence after i-th token. Since $y_w$ and $y_l$ are different only at one token, it is possible that $\log\pi_{\theta}(t_{i}^w, y^{> i}|x, y^{<i}) \approx \log\pi_{\theta}( t_{i}^l, y^{> i}|x, y^{<i})$, and $c\approx 0$, which causes the gradient to vanish. We empirically validate this phenomenon by plotting the training loss (Figure~\ref{subfig:training_loss}) and the gradient norm (Figure~\ref{subfig:gradient_norm}) during the training of DPO on the PTC dataset.

In addition to the weakening of the update signal resulting from the vanishing gradient, the small differences between $y_w$ and $y_l$ further lead to reduced probability of positive samples, as demonstrated in~\cite{pal2024smaug}. We also empirically validate this phenomenon by plotting the log probabilities for both positive and negative samples (see Figure~\ref{subfig:logps_chosen} and Figure~\ref{subfig:logps_rejected}). Our observations indicate that during training, although the log probabilities of negative samples decrease, the log probabilities of positive samples also decrease.

\section{The Comparison Between PPO and KTO}

Besides DPO, Proximal Policy Optimization (PPO) is also a widely used method to xx. However, is not usefull under our setting. We here conduct experiments comparing HiTEC-KTO with PPO. The PPO setting uses the same negative samples generated as described in Section \ref{sec:negative_sample}. Results are presented in Table \ref{tb:ppo}. We use Llama-Factory \footnote{\url{https://github.com/hiyouga/LLaMA-Factory/tree/main}} to first train a reward model and then conduct PPO.

\begin{table*}[h]
\centering
\fontsize{9}{10}\selectfont
\setlength{\tabcolsep}{2.5pt} % 调小列间距
\caption{Comparison Between PPO and KTO}
\begin{tabular}{@{}cccccc@{}}
\toprule
Dataset                      & Method    & \multicolumn{2}{c}{Qwen2.5-1.5B}                                            & \multicolumn{2}{c}{Qwen2.5-7B}                                              \\ \midrule
                             &           & F1 Name        & \begin{tabular}[c]{@{}c@{}}F1 Name\\ + Param.\end{tabular} & F1 Name        & \begin{tabular}[c]{@{}c@{}}F1 Name\\ + Param.\end{tabular} \\ \midrule
\multirow{2}{*}{Tool-Alpaca} & HiTEC-PPO & 7.01           & 1.23                                                       & 13.16          & 4.37                                                       \\
                             & HiTEC-KTO & \textbf{82.27} & \textbf{52.26}                                             & \textbf{87.63} & \textbf{59.19}                                             \\ \midrule
\multirow{2}{*}{Seal-Tools}  & HiTEC-PPO & 8.84           & 2.21                                                       & 17.68          & 8.15                                                       \\
                             & HiTEC-KTO & \textbf{96.99} & \textbf{89.75}                                             & \textbf{96.27} & \textbf{89.29}                                             \\ \bottomrule
\end{tabular}
\label{tb:ppo}
\end{table*}

As demosntrated in the table, there is a huge performance gap between HiTEC-PPO and HiTEC-KTO. This gap can be attributed to two well-known limitations of PPO: 1) poor generalization of the reward model – while the reward model performs well on the training dataset, it lacks generalization to unseen data, leading to suboptimal reinforcement learning. And 2) susceptibility to reward hacking, where the model learns to optimize for the reward function in unintended ways rather than genuinely improving task performance \cite{fu2025reward,gao2023scaling}.

A supporting example further illustrates this issue. Our trained reward model demonstrated high accuracy during both training and evaluation; however, it assigned a high score to the following output, despite the response containing formatting errors and irrelevant code snippets. 

This example indicates that the proposed generated dataset is not well-suited for PPO training. In contrast, KTO effectively integrates human demonstrations to refine the reward function, aligning it more closely with real-world preferences \cite{ethayarajh2024kto}. This alignment explains why KTO significantly outperforms PPO in our experiments and why it is a better fit for the current dataset.

\begin{tcolorbox}[colback=gray!10, colframe=gray!80, title=Example: The PPO Reward Model's High-Scoring of Model Responses]
\begin{verbatim}
Here is my response:

[
{
"name":"PublicHolidayNextPublicHolidays",
"arguments":{
"countryCode":"CN" // Replace with desired country code if different than GB
}
},

[
{
"name": "CountryCountryInfo",
"arguments":{"countryCode":"cn"}
}

# Add any additional functions here

]
]
<|file_sep|><|fim_prefix|>/README.md
import sys

sys.path.append("src")
from qwen_agent.agent import Agent
from qwen_<|fim_suffix|>tream=True,
stop=["

"],
)

agent = Agent(
llm=llm,
)

result = agent.run(user_message="Get me some jokes.")
print(result)<|fim_middle|>agent.tools.qwen_tools import *
from langchain.chat_models import ChatOpenAI

api_key_file_path = "/home/ubuntu/qwen-agent/.en
...
\end{verbatim}
\end{tcolorbox}